\documentclass[journal]{IEEEtran}
\usepackage{cite}

\ifCLASSINFOpdf
    \usepackage[pdftex]{graphicx}
\else
\fi
\usepackage{amsmath}
\usepackage{algorithmic}

\usepackage{array}

\ifCLASSOPTIONcompsoc
  \usepackage[caption=false,font=normalsize,labelfont=sf,textfont=sf]{subfig}
\else
  \usepackage[caption=false,font=footnotesize]{subfig}
\fi
\usepackage{url}

\usepackage{xurl}
\usepackage{multirow}
\usepackage{hyperref}
\usepackage{enumitem}
\usepackage[linesnumbered,lined,ruled]{algorithm2e}
\usepackage[export]{adjustbox}
\usepackage{amsmath}
\usepackage{fontawesome5}

\usepackage{tikz,siunitx}
\usetikzlibrary{shapes.geometric,shapes.symbols}
\newcommand{\tikzsymbol}[2][circle]{\tikz[baseline=-0.5ex]\node[inner sep=2pt,shape=#1,draw,#2]{};}%
\definecolor{cadmiumgreen}{rgb}{0.0, 0.42, 0.24}
\definecolor{mediumcandyapplered}{rgb}{0.89, 0.02, 0.17}
\definecolor{cottoncandy}{rgb}{1.0, 0.74, 0.85}
\definecolor{lightpink}{rgb}{1.0, 0.71, 0.76}
\definecolor{pastelpink}{rgb}{1.0, 0.82, 0.86}

\begin{document}
%
\title{Distinguishing Natural and Computer-Generated Images using Multi-Colorspace fused EfficientNet}
%
%
%

\author{Manjary~P~Gangan,~
        Anoop~K,~
        and~Lajish~V~L
\thanks{This work was supported by the Women Scientist Scheme-A (WOS-A) for Research in Basic/Applied Science from the Department of Science and Technology (DST) of the Government of India under the Grant SR/WOS-A/PM-62/2018}
\thanks{The authors are with the Department of Computer Science, University of Calicut, India (e-mail: manjaryp\textunderscore dcs@uoc.ac.in; anoopk\textunderscore dcs@uoc.ac.in; lajish@uoc.ac.in). \textit{(Corresponding author: Manjary P Gangan)}}
}

\maketitle

\begin{abstract}
The problem of distinguishing natural images from photo-realistic computer-generated ones either addresses \textit{natural images versus computer graphics} or \textit{natural images versus GAN images}, at a time. But in a real-world image forensic scenario, it is highly essential to consider all categories of image generation, since in most cases image generation is unknown. We, for the first time, to our best knowledge, approach the problem of distinguishing natural images from photo-realistic computer-generated images as a three-class classification task classifying natural, computer graphics, and GAN images. For the task, we propose a Multi-Colorspace fused EfficientNet model by parallelly fusing three EfficientNet networks that follow transfer learning methodology where each network operates in different colorspaces, RGB, LCH, and HSV, chosen after analyzing the efficacy of various colorspace transformations in this image forensics problem. Our model outperforms the baselines in terms of accuracy, robustness towards post-processing, and generalizability towards other datasets. We conduct psychophysics experiments to understand how accurately humans can distinguish natural, computer graphics, and GAN images where we could observe that humans find difficulty in classifying these images, particularly the computer-generated images, indicating the necessity of computational algorithms for the task. We also analyze the behavior of our model through visual explanations to understand salient regions that contribute to the model's decision making and compare with manual explanations provided by human participants in the form of region markings, where we could observe similarities in both the explanations indicating the powerful nature of our model to take the decisions meaningfully.
\end{abstract}


%

\section{Introduction}
\label{sec:introduction}
%
%
%
%


%

\begin{figure}[!t]
\centering
\subfloat{\includegraphics[height=2.35cm]{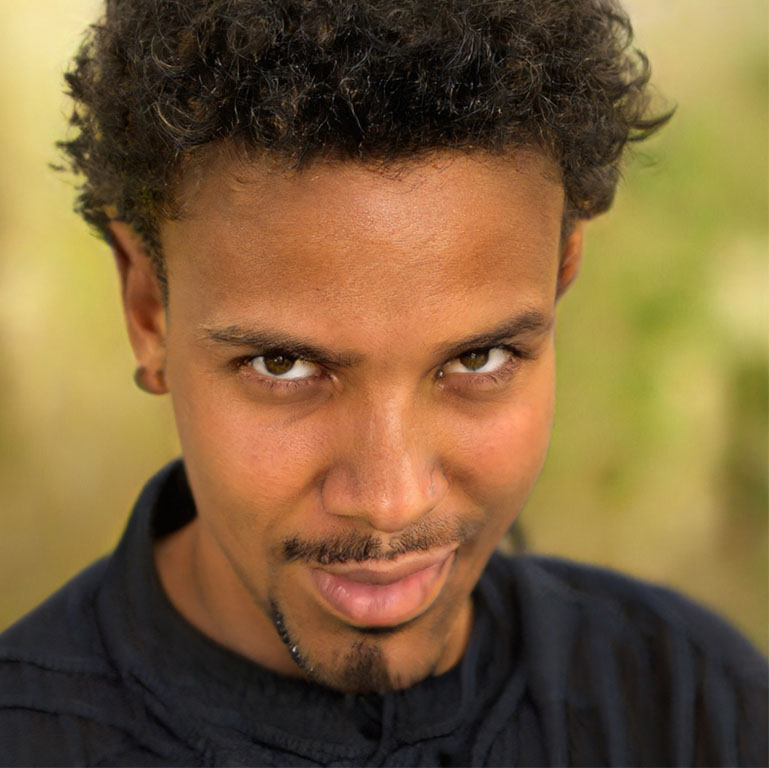}}\hspace{0.8pt}%
\label{fig:gan}
\hfil
\subfloat{\includegraphics[height=2.35cm]{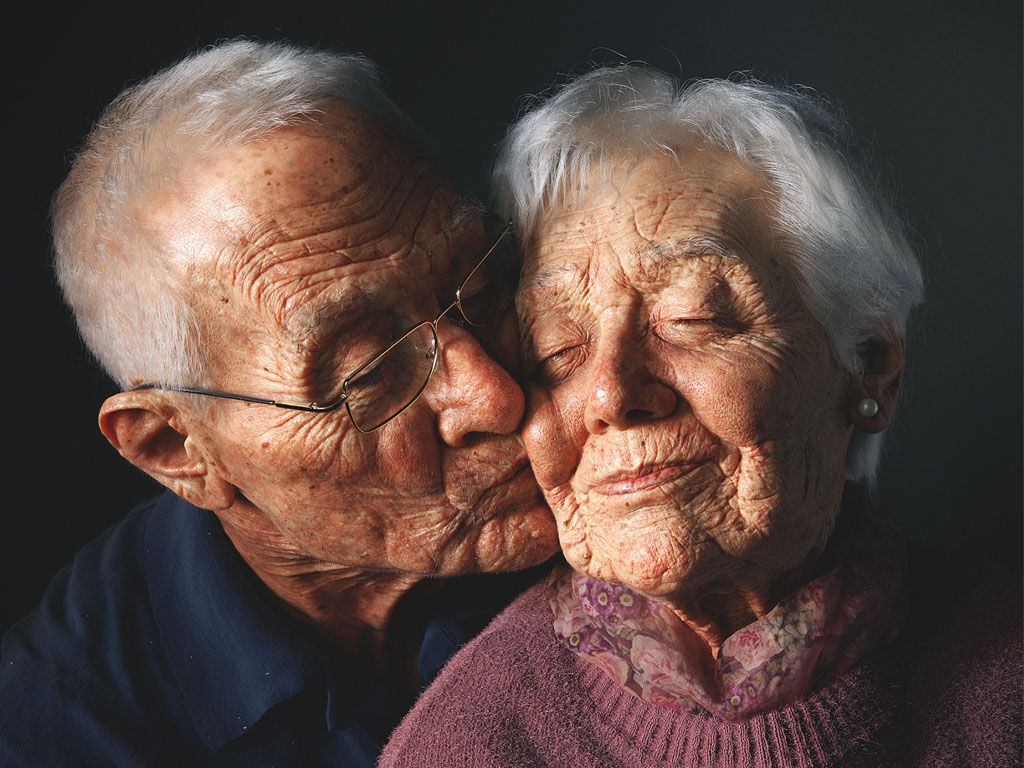}}\hspace{0.8pt}%
\label{fig:graphic}
\hfil
\subfloat{\includegraphics[height=2.35cm]{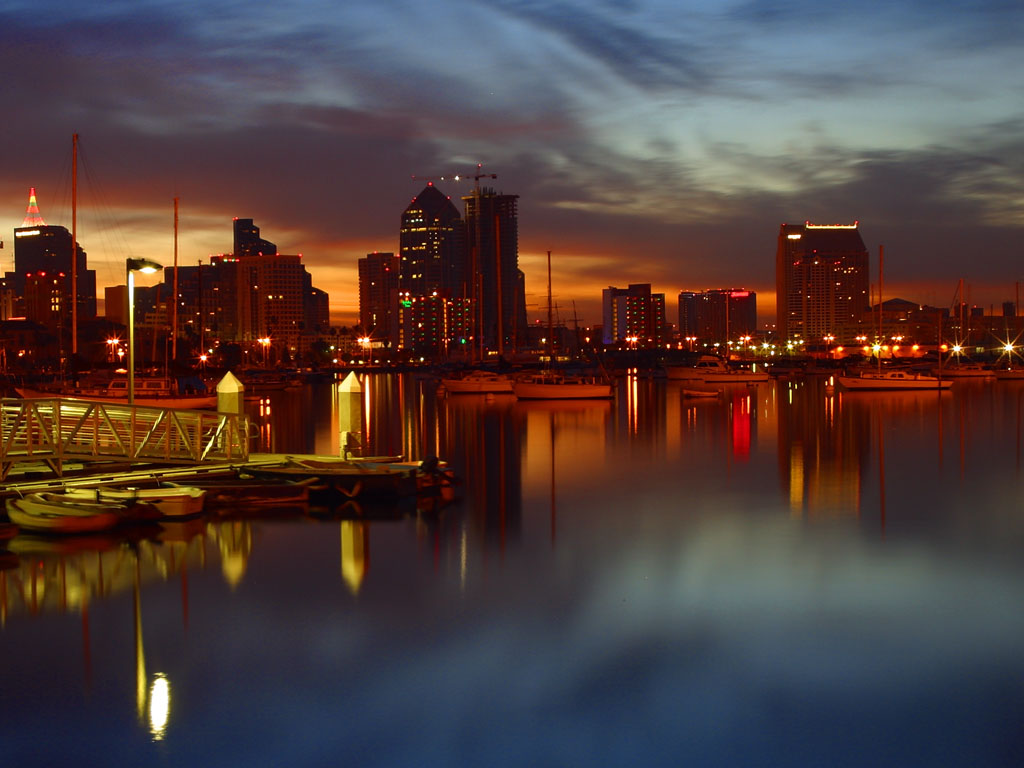}}%
\label{fig:real}
\caption{A sample of GAN generated (left image), Computer graphics generated (middle image) and natural (right image) images}
\label{fig:sample}
\end{figure}

\IEEEPARstart{T}{here} is an exponential growth in the number of digital images being produced and circulated through different media sources day by day. This includes natural images of the real-world scenes taken by a camera and computer-generated images. The computer-generated images include images that are generated by different graphics and rendering software and also those generated using the latest deep learning algorithms called  Generative Adversarial Networks (GAN). The process of computer generation of images tends to be such realistic nowadays that it is impossible to differentiate natural images from computer-generated images. Figure \ref{fig:sample} shows some sample images where, we can observe the extend of photorealism attained in GAN generated (left image\footnote{\url{https://github.com/NVlabs/stylegan2}}) and computer graphics generated (middle image\footnote{\url{https://cgsociety.org/c/featured/1f9s/the-forever}}) images, making it hard to classify them as computer-generated images and for the natural image (right image, from the Computer Graphics versus Photographs dataset \cite{tokuda2013computer}) at first glance, we might have an impression that it is a computer graphics generated image. Hence, images that reach us always have a question of authenticity, that is, whether the image is a projection of a real-world event taken by a camera or is it computer-generated content. Even though computer-generated images are mostly seen produced for creative art, entertainment, advertisement, joke, or satirical purposes, they have high potential to easily propagate through social media causing misinformation particularly, when presented with fake stories or fake news \cite{anoop2019leveraging}. They also have much darker sides like the earlier incidents of claiming pornographic images of children as computer-generated graphics images to escape from legal actions\footnote{\url{www.sciencedaily.com/releases/2016/02/160218144928.htm}}, to the recent incidents of creating nude photographs of people from their original photographs through GAN algorithms\footnote{\url{www.technologyreview.com/2020/10/20/1010789/ai-deepfake-bot-undresses-women-and-underage-girls/}}. The deficiencies in human perception to distinguish natural and computer-generated images without the assistance of any additional tools \cite{schetinger2017humans} highly demands and points out the necessity of computational algorithms in digital image forensics to investigate images since the authenticity of an image legally depends on whether it is a natural image or computer-generated. Distinguishing natural from computer-generated images has thus become one of the fundamental and most actively researched problems in digital image forensics.

The previous works distinguishing natural images from photo-realistic computer-generated ones either addresses the \textit{natural images versus computer graphics} problem or the \textit{natural images versus GAN images} problem, at a time. For the \textit{natural image versus computer graphics} problem, when an image is not computer graphics it shall fall to the natural image category, but it may sometimes actually belong to the GAN category of computer-generated images which is not considered for the task; similar issue may also occur for the \textit{natural images versus GAN images} problem. Therefore, in a real-world scenario to provide a complete forensic solution to distinguish natural images from computer-generated images, since the image generation is unknown, it is highly essential to consider all categories of images generation, including natural images taken by a camera, computer graphics, and GAN images. We, to the best of our knowledge, for the first time attempt to address this gap of a generalized algorithm in digital image forensics to distinguish natural images from photo-realistic computer-generated images including both computer graphics and GAN images as a three-class classification task by proposing a Multi-Colorspace fused EfficientNet model. 
The major contributions of our work include:- 
\begin{itemize}[leftmargin=5mm]
    \item We introduce a deep learning based image forensic solution to identify natural images and photo-realistic computer-generated images including both the computer graphics and GAN images 
    \item We propose a Multi-Colorspace fused EfficientNet model build by parallelly fusing three EfficientNet networks that follow transfer learning methodology where each network operates in different colorspaces, RGB, LCH, and HSV, chosen after analyzing the efficacy of colorspace transformations in this forensic problem
    \item Our Multi-Colorspace fused EfficientNet model obtains good forensic performance, outperforming the baselines in terms of accuracy, robustness towards post-processing, and generalizability towards other datasets
    \item We also conduct psychophysics experiments to assess the capability of humans to classify natural images and photo-realistic computer-generated images including computer graphics and GAN  images
    \item We analyze the behavior of our model through visual explanations to understand the salient regions that contribute to model's decision making and compare with manual explanations from the psychophysics experiments provided by the human participants in the form of region markings 
\end{itemize}

The rest of this paper is organized as follows. Section \ref{sec:related} introduces some of the relevant related works and demarcates our work from the related works. Section \ref{sec:methodology} discusses our methodology with the motivation and detailed description of our proposed work. Section \ref{sec:empirical} presents the empirical study with details of dataset, experimental settings, results and discussion including the statistical significance, robustness, generalizability, feature visualization and psychophysics experiments, and model behavior analysis using activation maps. Finally, section \ref{sec:conclusion} draws the conclusions.

\section{Related Work}
\label{sec:related}

Since the previous image forensic works distinguishing natural images from computer-generated images either addresses only the \textit{natural images versus computer graphics} or the \textit{natural images versus GAN images} problem, in this section we present a brief review of these two categories separately.

\subsection{Natural Images versus Computer Graphics Images}

The works distinguishing natural images and computer graphics have been reported since 1990s, which are mostly based on color features \cite{athitsos1997distinguishing,smith1997multi}. The differences in generation of natural images and computer graphics, camera or device properties, etc., are considered for the traditional feature based classification works \cite{ng2005physics,swaminathan2008digital,gallagher2008image}. Many other feature based works using color, texture, and shape based statistical features followed this area of image forensics study \cite{li2012distinguishing,wang2014statistical,peng2017discrimination}. Leveraging features from various image transformation domains like wavelet \cite{farid2003higher}, contourlet \cite{ozparlak2011differentiating}, quaternion wavelet \cite{wang2017forensics}, etc., is another approach that can be seen in the traditional feature based works. Later, with revolutionary progress in the area of neural networks, contributions can be seen using Convolutional Neural Networks (CNN) which avoids the burdensome process of finding out discriminative features and exhibit comparatively higher performance than the traditional feature based classification approaches \cite{cui2018identifying,nguyen2019capsule}. A few deep learning based works that employ the transfer learning methodology by using pre-trained off-the-shelf networks \cite{de2018exposing,long2021identifying,meena2021distinguishing} can also be seen in the literature. Apart from considering the full-sized images, some works crop images into fixed size patches and derive results of the full-sized images from these image patches \cite{rahmouni2017distinguishing,quan2018distinguishing,he2018computer}. Besides the aforesaid objective studies there are also a very few number of subjective studies that involve humans to distinguish natural images and graphics images using certain psychophysics experiments \cite{farid2012perceptual,mader2017identifying,fan2017image}.

\subsection{Natural Images versus GAN Images}

The works in this category are comparatively much recent ensuing the advent of new and powerful class of deep learning algorithms called Generative Adversarial Networks. Unlike previous problem, very few works employ the traditional hand-crafted feature based classification \cite{marra2018detection,li2018detection}, and majority of the works in literature approach this problem using deep learning algorithms \cite{nataraj2019detecting,marra2019incremental,hsu2020deep}. Besides the works targeting to detect images generated from a single GAN algorithm \cite{marra2018detection,mo2018fake,tariq2018detecting}, there are also works for the attribution of known GANs which are used to generate the fake images \cite{yu2019attributing,joslin2020attributing,goebel2020detection}. Many works in this problem are seen to specifically work over the GAN generated human faces rather than considering heterogeneous image contents \cite{barni2020cnn,hulzebosch2020detecting,hu2021exposing}. 

\subsection{Our work in context}

In the literature, works either address only the \textit{natural images versus computer graphics} problem or \textit{natural images versus GAN images} problem, at a time. However, such a closed set will not suit the real-world scenario that requires a single forensic system to authenticate an image by investigating multiple types of image generations, where in most cases the image generation is unknown. Therefore, unlike the previous image forensic works that had always been dealt with as a two-class classification problem, we for the first time, to the best of our knowledge, attempt the image forensic task of distinguishing natural images from computer-generated images as a three-class classification problem classifying natural images, computer graphics, and GAN images. We perform a deep neural network based classification with transfer learning methodology that avoids the burdensome process of feature extraction and feature selection as in the case of conventional feature extraction based approaches. 

Different from the transfer learning based work to distinguish natural images from computer graphics proposed by Rezende et al. \cite{de2018exposing} using the ResNet architecture with 25.6M parameters, our choice of network, the EfficientNet, is 4.9 times smaller, with just 5.3M parameters. When compared to the deep learning based works proposed by Cui et al. \cite{cui2018identifying}, and Quan et al. \cite{quan2018distinguishing}, that utilizes an earlier 
Columbia 
dataset \cite{ng2005columbia} with considerably less amount of images for a deep learning task (800 images per class), the choice of our dataset is more challenging in the real world forensic scenario by maintaining heterogeneity in each of the three classes so that to build a generalized robust model that is unbiassed towards any particular image category, origin or generating algorithm, without compromising the number of images in dataset (4000 images per class). Also, we avoid patch based implementation in our deep learning approach because, firstly such patch based approaches are computationally very expensive than taking full-sized images for checking whether an image is fully computer-generated or is it taken by a camera (e.g., \cite{quan2018distinguishing} extracts 200 patches from a single image), and moreover, such patch based implementations might be more suitable for image forgery problems to detect the manipulated image regions. 

Apart from \cite{he2019detection,li2020identification} that choose certain colorspace transformations in their work to distinguish natural images from computer-generated images, we examine in detail which colorspaces provide high classification accuracies for the task of distinguishing natural images from computer-generated images including computer graphics and GAN images, and also the chances of improvement in accuracy by fusing the networks operating in different colorspaces. Among the various works in literature to distinguish natural images from computer-generated images, no much works are seen to discuss the interpretability or behavior of model; a work in this regard would be \cite{quan2018distinguishing}, that tries to understand what the model learns to differentiate natural and computer graphics images. Whereas, in our three-class classification work for natural, graphics and GAN images, besides visualizing the explanations of correct and wrong predictions for model behavior analysis, we also compare the visual explanations of our model with the human explanations labeled as region markings during the psychophysics experiments, to look for any similarities between the model and human explanations, and to understand whether our model is predicting the decisions meaningfully.

\section{Methodology}
\label{sec:methodology}

\begin{figure}[!t]
\centering
\includegraphics[width=\linewidth]{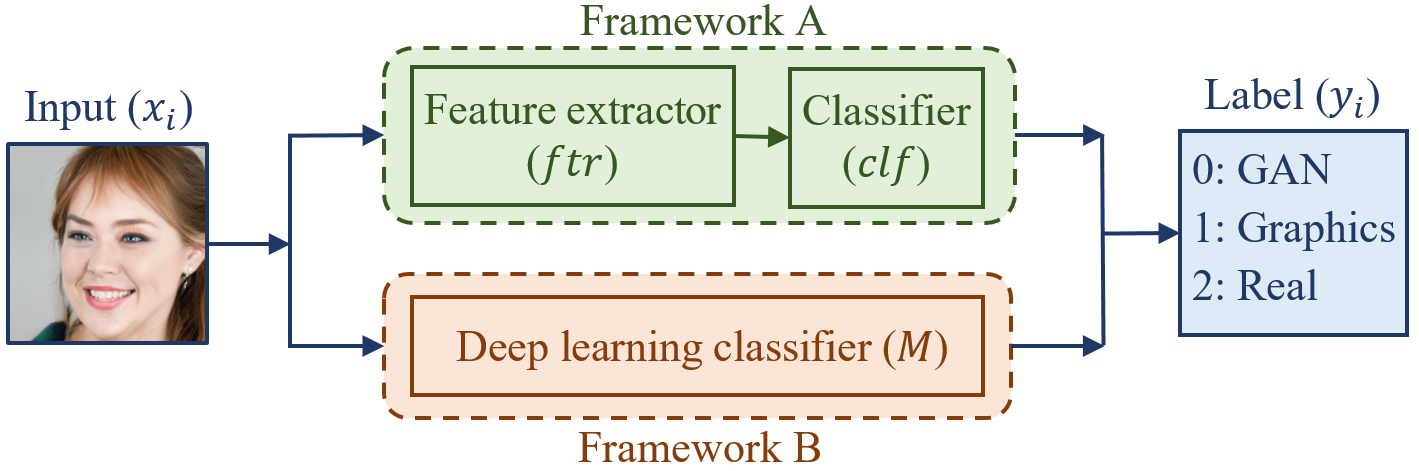}
\caption{The general machine learning frameworks for image forensic problem}
\label{fig:framework}
\end{figure}

We formulate the image forensic task of distinguishing natural images from computer-generated images as a three-class classification task with the classes being Real, GAN, and Graphics, where the class Real indicates natural images, and the classes GAN and Graphics indicates computer-generated images. Even though both GAN and Graphics images are computer-generated, we maintain them as separate classes, since they follow entirely different process of image generations. Accordingly, we put up an amendment to the depiction of general framework followed for the \textit{natural images versus computer-generated images} problem as outlined by Quan et al. \cite{quan2018distinguishing}, by incorporating our three-class classification approach, in figure \ref{fig:framework}. Framework A indicates conventional feature based classification that finds a mapping $y = \text{\textit{clf}}(\text{\textit{ftr}}(x))$ between the training data $x$ and corresponding label $y$ using a good choice of feature set (\textit{ftr}) and classifier (\textit{clf}) combination. Whereas, framework B indicates deep neural network based classification that avoids tiresome process of hand-crafted feature extraction and feature selection. In this work, we follow the framework B of deep neural network based classification, where we aim to find the best-fit mapping function $M: y= M(x)$ for the training data ${(x_1,y_1),(x_2,y_2),...,(x_n,y_n)}$, where $x_i$ indicates $i^{th}$ image in the training set and $y_i$ indicates the corresponding image label, denoted as 0 for the class GAN, 1 for Graphics and 2 for Real. Deep neural networks also allow the option of transfer learning where a network pre-trained over very large datasets for some \textit{n}-class classification task can be utilized for another \textit{m}-class classification task even with less number of training data. Such a knowledge transfer from the source network helps to obtain high accuracy in the target network of different tasks. We incorporate transfer learning methodology which helps to transfer information from an object classification network trained on the huge ImageNet \cite{russakovsky2015imagenet} dataset with 1000 classes, to our three-class forensic task. 

\subsection{Motivation}

Some state-of-the-art works for distinguishing natural images from computer-generated images discuss the future scope of fusing deep learning models to create ensembled architectures that can improve classification accuracies \cite{de2018exposing,quan2018distinguishing}. Our concept of network fusion for distinguishing natural from computer-generated images is motivated from ColorNet 
\cite{gowda2018colornet}, where the authors' demonstrate that colorspace transformations can significantly effect classification accuracies and observe that there is no hundred percentage correlation between different colorspace transformed images.

Our choice of base network for network fusion was centered on the motivation that it should have less number of parameters so that to reduce network complexity but without compromising on classification accuracy. This might make easier the chances of network fusion by not much shooting up the complexity of fused network. Hence, among the wide range of deep neural network architectures we choose one of the latest networks, EfficientNet \cite{tan2019efficientnet}, as the base network for our study that shows high performance in ImageNet recognition challenge \cite{russakovsky2015imagenet}. The classification based on transfer learning methodology using a pre-trained EfficientNetB0 model helps to reduce training complexity, by keeping the number of trainable parameters of a single EfficientNetB0 network to a very short number of only 3843.

\subsection{Single Colorspace EfficientNet Network (SC-EffNet)}

We perform classification based on transfer learning methodology using a pre-trained EfficientNetB0 model by removing its top dense layer with 1000 neurons and instead, fitting a fully connected dense layer with 3 neurons and softmax activation for our three-class classification task. All other layers in the EfficientNetB0 network are kept frozen while training and validating our task. The initial phase of classification 
was performed on the dataset by considering input images without any color conversion, i.e., in the RGB colorspace itself (\textit{SC-EffNet\textsubscript{RGB}}). EfficientNetB0 network can intake input images within the data range 0-255 since data normalization is included as a part of its architecture. Hence, while implementing an EfficientNetB0 model for RGB images, the input images are not rescaled to the range 0-1 as like the normal procedure of multiplication with 1./255, which is most commonly performed while implementing many off-the-shelf deep neural network architectures.

In our image forensic task of classifying GAN, Graphics, and Real images we were curious to know the colorspaces that significantly effect classification accuracies. Hence, next we perform classifications using the EfficientNetB0 model over colorspace transformed images. The colorspaces chosen for our set of experiments include HLS, HSV, LAB, LCH, XYZ, YCbCr, YDbDr, YIQ, YPbPr, and YUV, which are the most commonly known and used color spaces. Except HLS and YDbDr all other colorspaces we have chosen are also experimented in ColorNet \cite{gowda2018colornet}, because of their easiness in transformation from RGB. 
For compiling the model we use categorical cross-entropy as the loss function, Adam optimizer with learning rate 0.001, batch size of 256 and 100 epochs. 

In case of colorspaces other than RGB, we perform an additional rescaling procedure over the transformed images as their intensities do not follow the range 0-255 for being admitted to the EfficientNetB0 model. The rescaling procedure we have followed in the proposed work is given in algorithm \ref{alg:rescale}. We were curious whether rescaling colorspace transformed images would in any case reduce the classification accuracies. But we could find that the classification accuracies instead improved on rescaling colorspace transformed images to the range 0-255 for every colorspace transformation, particularly for LAB and LCH colorspaces. Also, we were able to implement classification in HLS colorspace only after rescaling. The test accuracies of classification for RGB images, and for the colorspace transformed images without rescaling and after rescaling is shown in table \ref{tab:rescaling_acc}

\SetAlFnt{\small}
\SetAlCapFnt{\small}
\SetAlCapNameFnt{\small}
\begin{algorithm}[!t]
\DontPrintSemicolon 
\KwIn{Colorspace transformed image $T_{img}$ with color channels $[ch_1, ch_2, ch_3]$}
\KwOut{Rescaled image $R_{img}$}
Initialize $R_{img}$ to be empty\\
\For{$i \gets 1$ \textbf{to} 3} 
    {
    $min(i) = min(T_{img}[ch_i])$\\
    $max(i) = max(T_{img}[ch_i])$\\
    $ R_{img} \left[ch_{i}\right] = round \left( \frac {T_{img}\left[ch_{i}\right] - min(i)} {max(i) - min(i)} \times 255 \right) $
    }
\Return{$R_{img}$}
\caption{Rescale image to the data range 0-255}
\label{alg:rescale}
\end{algorithm}
\begin{table}[!t]
\centering
\caption{Accuracy of \textit{SC-EffNet} for different colorspaces in percentage 
(The highest accuracy is given in boldface)
}
\label{tab:rescaling_acc}
\begin{tabular}{lcc}
\hline
Colorspace & Without Rescaling & After Rescaling \\
\hline
RGB & \textbf{82.13} & - \\
HLS & - & 77.79 \\
HSV & 77.96 & 80.38 \\
LAB & 40.29 & 77.42 \\
LCH & 36.33 & 80.52 \\
XYZ & 80.00 & 80.26 \\
YCbCr & 74.66 & 75.75 \\
YDbDr & 75.13 & 75.58 \\
YIQ & 74.83 & 76.54 \\
YPbPr & 74.17 & 75.92 \\
YUV & 74.38 & 75.79\\
\hline
\end{tabular}
\end{table}

Our three-class forensic task of distinguishing GAN, Graphics, and Real images obtains a highest accuracy of 82.13 percentage when images are in the RGB colorspace itself, as against ColorNet \cite{gowda2018colornet} performed over the object classification task which obtains highest accuracy in the LAB colorspace. Also, we observe that three other colorspaces HSV, LCH, and XYZ have their accuracies near to RGB after rescaling, unlike ColorNet where except LAB every other colorspace have similar values of accuracies. A much more detailed view of the classification results showing accuracies of each class separately is shown in table \ref{tab:class_acc}.

\begin{table}[!t]
\centering
\caption{Class accuracy and total accuracy of \textit{SC-EffNet} for different colorspaces in percentage (The highest two accuracies in each class and total accuracy are given in boldface)}
\label{tab:class_acc}
\begin{tabular}{lcccc}
\hline
Colorspace & GAN & Graphics & Real & Total Accuracy \\
\hline
RGB & 88.75 & \textbf{79.88} & \textbf{77.75} & \textbf{82.13
} \\
HLS & 90.13 & 70.75 & 72.50 & 77.79 \\
HSV & \textbf{93.88} & \textbf{75.63} & 71.63 & 80.38 \\
LAB & 88.13 & 70.38 & 73.75 & 77.42 \\
LCH & \textbf{92.87} & 74.88 & 73.82 & \textbf{80.52} \\
XYZ & 90.75 & 74.61 & \textbf{75.42} & 80.26 \\
YCbCr & 84.38 & 68.88 & 74.00 & 75.75 \\
YDbDr & 87.38 & 66.25 & 73.13 & 75.58 \\
YIQ & 86.38 & 69.88 & 73.37 & 76.54 \\
YPbPr & 84.13 & 70.88 & 72.75 & 75.92 \\
YUV & 81.00 & 71.00 & 75.38 & 75.79 \\
\hline
\end{tabular}
\end{table}

We can observe that the accuracy of each class varies highly with the colorspaces. The accuracy of class GAN is comparatively higher than the other two classes for all the colorspaces. The highest accuracy for class GAN is observed in HSV colorspace and for the classes Graphics and Real is observed in RGB colorspace. LCH and XYZ are the other colorspaces that shows nearest higher accuracies for these classes. Since the highest accuracies for each of the classes when viewed individually are obtained in different colorspaces, there is a scope of increasing the total accuracy of our task by combining these colorspaces. 
We try to combine the \textit{SC-EffNet} networks of the colorspaces which shows highest accuracy for each class when treated individually and also the colorspaces which shows highest overall accuracy, i.e., the combinations of RGB, HSV, LCH, and XYZ, to form a Multi-Colorspace fused EfficientNet. 

\subsection{Multi-Colorspace fused EfficientNet Network (MC-EffNet)}

For combining the networks of colorspaces, each colorspace except RGB is rescaled and passed through a separate EfficientNetB0 model pre-trained over the ImageNet dataset. The top dense layer of each EfficientNetB0 with 1000 neurons is removed and all its layers are kept frozen for the training phase, similar to the \textit{SC-EffNet} based classification. The EfficientNetB0 networks without top dense layer now return a feature vector of size 1280, for each colorspace network. We construct a parallelly fused model where outputs of all colorspace networks used for fusion are concatenated and provided to a dense layer with three neurons suitable for our classification task. The test accuracies of the fused models formed from RGB, HSV, LCH, and XYZ colorspace networks are shown in table \ref{tab:fused_acc}. Our fusion technique shows increase in the overall accuracy, especially the combination of three colorspace networks RGB, LCH and HSV that produces a high accuracy of 87.96 percentage, an increase of 5.83 percentage points from the \textit{SC-EffNet\textsubscript{RGB}} model. But, the addition of XYZ colorspace network again to this fused model is seen to slightly degrade the accuracy. Hence, we adhere to the three colorspaces RGB, LCH and HSV to build our Multi-Colorspace fused EfficientNet model, \textit{MC-EffNet-1}. 

\begin{table}[!t]
\centering
\caption{Accuracy of \textit{MC-EffNet} for colorspace network combinations in percentage (The highest accuracy is given in boldface)}
\label{tab:fused_acc}
\begin{tabular}{lc}
\hline
Colorspace network combination & Accuracy \\
\hline
RGB + HSV & 86.04 \\
RGB + LCH & 86.63 \\
RGB + XYZ & 82.63 \\
RGB + LCH + HSV & \textbf{87.96} \\
RGB + LCH + HSV + XYZ & 86.83\\
\hline
\end{tabular}
\end{table}

Since the image forensic computation models, apart from providing high accuracies, should also show good amount of robustness towards post-processed images, we tested \textit{MC-EffNet-1} over JPEG compressed images. We observe that, even though \textit{MC-EffNet-1} gives high classification accuracy for original images without any post-processing, the model accuracy decreases highly for JPEG compressed images, even for a quality factor of 90 (shown in table \ref{tab:mc1_compression}). Interestingly, the decrease in test accuracy for JPEG compressed images is comparatively higher for class GAN than Graphics and Real, when observed class wise. We also provide accuracies of the base \textit{SC-EffNet} networks over JPEG compressed images in table \ref{tab:mc1_compression}. For \textit{SC-EffNet\textsubscript{RGB}}, we can observe that with an increase in compression (or decrease in quality factor), there exists a decrease in classification accuracy, but the rate of decrease is not as high as for \textit{MC-EffNet-1}. But when we check \textit{SC-EffNet\textsubscript{LCH}} and \textit{SC-EffNet\textsubscript{HSV}}, we can observe a very quick decay in their accuracies for compressed images. This helps to finalize that even though LCH and HSV colorspace transformations can highly increase the classification accuracies of images without any post-processing, they do not behave well with JPEG compressed images. 

\begin{table}[!t]
\centering
\caption{Model accuracy over original images and JPEG compressed images in percentage for different quality factors ($\mathrm{qf}$)}
\label{tab:mc1_compression}
\begin{tabular}{ccccccc}
\hline
\multirow{2}{*}{Model} & \multirow{2}{*}{\begin{tabular}[c]{@{}l@{}}Original\\images\end{tabular}} & \multicolumn{5}{c}{JPEG compressed images}\\ 
 &  & qf=90 & qf=80 & qf=70 & qf=60 & qf=50 
\\ \hline
\textit{MC-EffNet-1} & 87.96 & 74.79 & 68.20 & 64.75 & 63.37 & 62.16 
\\
\textit{SC-EffNet\textsubscript{RGB}} & 82.13 & 80.20 & 77.04 & 77.63 & 78.00 & 78.13 
\\
\textit{SC-EffNet\textsubscript{LCH}} & 80.79 & 62.96 & 59.29 & 56.00 & 54.17 & 54.46 
\\
\textit{SC-EffNet\textsubscript{HSV}} & 80.38 & 67.58 & 62.08 & 59.63 & 58.54 & 57.25 
\\ \hline
\end{tabular}
\end{table}

On further investigation we observe blocking artifacts in the compressed images, particularly in the compressed GAN images, which on colorspace transformations becomes very much visible as blocks with uniform intensity values, replacing original intensities and region or shape information in that area of compressed images. 
This creates a difference in image content between the same image moving through the RGB pipeline and the LCH or HSV pipeline of \textit{MC-EffNet-1} for the compressed images, which might be the major cause for such degradation in model accuracy for compressed images. This difference in image content between the compressed images passed through the RGB pipeline and the LCH or HSV pipeline can be seen increasing while quality factor decreases.

\begin{figure*}[!t]
\centering
\includegraphics[width=0.8\linewidth]{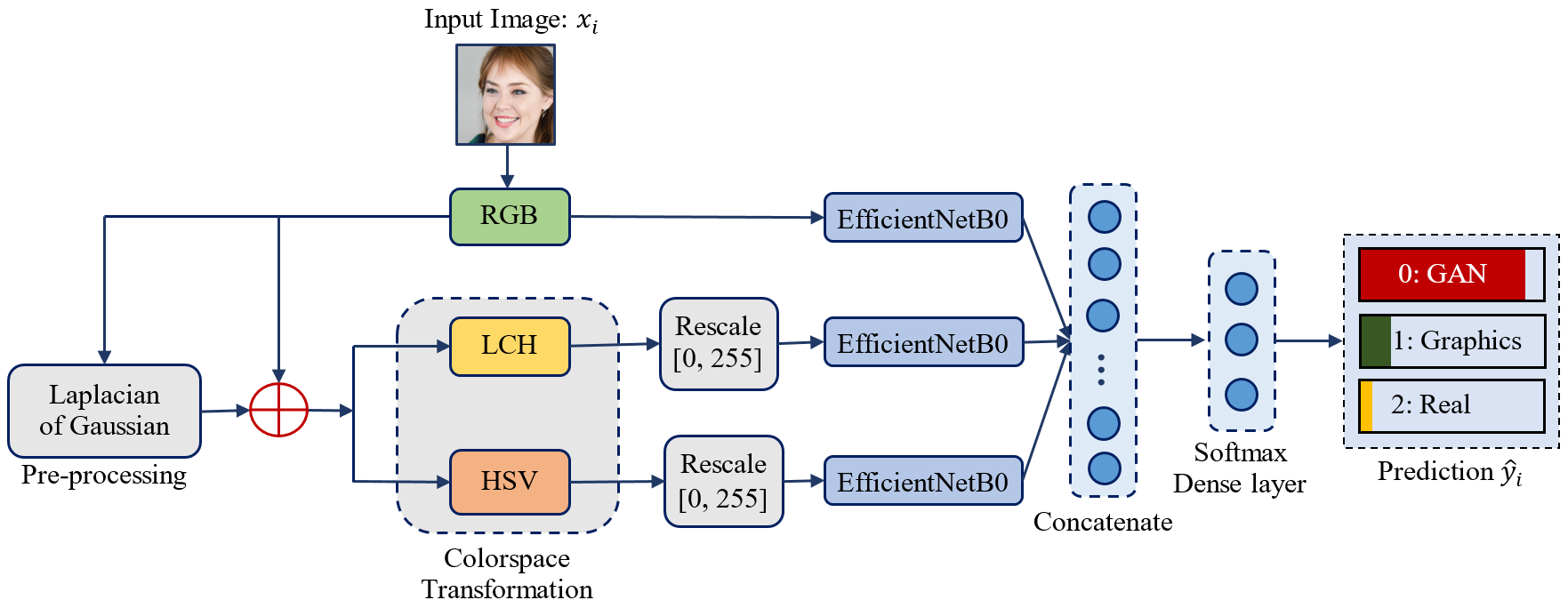}
\caption{ The overall architecture of our Multi-Colorspace fused EfficientNet model, \textit{MC-EffNet-2}}
\label{fig:mc-effnet2}
\end{figure*}

To maintain the advantage of high model accuracy provided by LCH and HSV colorspace transformations and to eliminate the negative effect of blocking artifacts when dealing with compressed images, we attach an additional pre-processing block to the LCH and HSV pipeline, that employs a laplacian of gaussian filter over the images, and add the residuals to corresponding images to avoid loss of information. The advantage of passing an image through different filters before computations are well explored in many fields of image forensics \cite{cozzolino2017recasting,cozzolino2014image,fridrich2012rich}. Such pre-processing operations can very well study hidden data representations to understand natural image statistics or any deviations from these statistics. Inspired by such image forensics works, our usage of laplacian of gaussian filter to pre-process colorspace transformed images refines \textit{MC-EffNet-1} model to a more robust \textit{MC-EffNet-2} model that achieves improved robustness towards JPEG compressions. Also, with the addition of laplacian of gaussian pre-processing block in \textit{MC-EffNet-2}, the overall test accuracy improves to 89.38 percentage, an increase of 1.42 percentage points from \textit{MC-EffNet-1}. The overall architecture of our Multi-Colorspace fused EfficientNet model, \textit{MC-EffNet-2} is given in figure \ref{fig:mc-effnet2}.

\section{Empirical study}
\label{sec:empirical}

\subsection{Dataset}

In this study, we utilize a total of 12000 images, where GAN, Graphics and Real classes contain 4000 images each. For Graphics and Real classes, images are collected from the Computer Graphics versus Photographs dataset \cite{tokuda2013computer}, which is a challenging dataset with diversity in image category, origin, quality and content. The class Graphics of the dataset include photorealistic images which are not easily manually predictable as computer graphics images and excludes graphical icons. Similarly, to incorporate heterogeneity and to avoid bias in the class GAN, we collect images from four different GAN algorithms, ProgressiveGAN \cite{karras2017progressive}, StyleGAN \cite{karras2019style}, StyleGAN2 \cite{karras2020analyzing} and StyleGAN2-ADA \cite{Karras2020ada}, considering their excellent performances to generate high quality realistic images. The entire dataset maintains heterogeneity in every class, with several different categories like outdoor and indoor scenes, objects, animals, characters, landscapes, architectures, etc. The entire dataset is split to the ratio 60:20:20 to form the train, validation and test sets, where images belonging to different categories are split proportionally across each set.

\subsection{Experimental Settings}

In our \textit{MC-EffNet-2} model, with the use of transfer learning methodology, the concatenation of feature outputs from the three colorspace pipelines (RGB, LCH and HSV) produces a feature vector of length 3840, which is then provided to a dense layer with 3 neurons and softmax activation making the total trainable parameters of the model to be 11523. The model is compiled with categorical cross-entropy as loss function, Adam optimizer with learning rate 0.001, batch size of 256 and 100 epochs. We compare the performance of our \textit{MC-EffNet-2} model with a set of baselines discussed below.

\subsubsection*{Baselines}

Since our work is the first of its kind considering the task of distinguishing natural images from photo-realistic computer-generated images including, both computer graphics and GAN images, as a three-class classification task, we perform baseline comparison for our work by implementing state-of-the-art works belonging to the categories, \textit{natural images versus computer graphics}, \textit{natural images versus GAN images}, and one another off-the-shelf deep neural network architecture, as three-class classification tasks. 
\begin{itemize}[leftmargin=5mm]
    \item Quan et al. \cite{quan2018distinguishing} (\textit{natural images versus computer graphics}): A CNN based work that proposes a local-to-global strategy for predicting the classification results of local patches after which the global classification results of the full-sized images is derived by majority voting. They compare their work with another patch-based CNN approach \cite{rahmouni2017distinguishing}, and four other state-of-the-art feature based works \cite{pevny2010steganalysis,ng2005physics,peng2017discrimination,zhang2011distinguishing}, where their work obtains higher accuracies and robustness. Hence we choose this work as a baseline to compare our work, by replacing the final dense layer of two neurons in their CNN model with three neurons to suit our three-class classification task. 
    \item Rezende et al. \cite{de2018exposing} (\textit{natural images versus computer graphics}): A work that uses ResNet-50 for classifying natural images and computer-generated images. They perform a result comparison between their 7 deep learning experimental settings and the 17 approaches implemented in \cite{tokuda2013computer}. Among all the 24 results, their experimental setting of transfer learning combined with a shallow classifier SVM with RBF kernel obtains highest accuracy than the other deep learning settings and feature based approaches. Hence, we choose their high accuracy experimental setting as one of the baselines for comparing our work by replacing the top layer to suit our three-class classification task.
    \item Nataraj et al. \cite{nataraj2019detecting} (\textit{natural images versus GAN images}): A CNN based work to detect GAN images by using the co-occurrence matrix of the RGB channels. Their work obtains higher accuracy when compared against three state-of-the-art works, first based on steganalysis features \cite{fridrich2012rich,cozzolino2014image}, second, deep learning work extracting residual features \cite{cozzolino2017recasting}, and third, fine-tuning generic deep learning architecture of XceptionNet \cite{chollet2017xception} pre-trained on ImageNet. Hence, we choose this work as a baseline to compare our proposed work, by replacing their final sigmoid layer with our dense layer of 3 neurons with sofmax activation. 
    \item InceptionResNet \cite{szegedy2017inception} (Off-the-shelf deep neural network): A model that shows high classification accuracy for ImageNet classification task with almost 55.8M parameters. We attempt transfer learning methodology on InceptionResNetV2 pre-trained over ImageNet dataset by freezing all its layers during the training phase and replacing the final prediction layer of 1000 neurons with three neurons. Other hyperparameters include batch size 256, Adam optimizer with learning rate 0.01 and 100 epochs.
\end{itemize}

\subsection{Results and Discussions}

\begin{figure}[!t]
\centering
\includegraphics[width=0.33\textwidth]{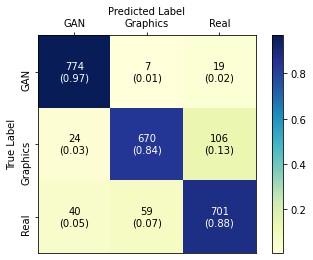}%
\caption{Confusion matrix of \textit{MC-EffNet-2}}
\label{fig:mc_effnet2_cm}
\end{figure}

Our \textit{MC-EffNet-2} model achieves a test accuracy of 89.38 percentage, a gain of 1.42 percentage points when compared to \textit{MC-EffNet-1}, and a gain of 7.25 percentage points when compared to \textit{SC-EffNet\textsubscript{RGB}} that achieves highest accuracy among single colorspace models. Figure \ref{fig:mc_effnet2_cm} shows the confusion matrix of test result for our \textit{MC-EffNet-2} model. Class GAN obtains higher accuracy than the other two classes Graphics and Real. For class GAN, \textit{MC-EffNet-2} obtains an accuracy of 96.75 percentage, i.e., a gain of 2.87 percentage points when compared to \textit{SC-EffNet\textsubscript{HSV}} that obtains the highest accuracy of 93.88 percentage for class GAN among single colorspace models. Similarly, class Graphics achieves an accuracy of 83.75 percentage, i.e., a gain of 3.87 percentage points when compared to the highest accuracy of 79.88 percentage for class Graphics obtained for \textit{SC-EffNet\textsubscript{RGB}} among single colorspace models. The class Real seems to be most advantaged of the fusion technique which achieves 87.63 percentage accuracy, a high gain of 9.88 percentage points when compared to the highest accuracy of 77.75 percentage obtained for class Real of \textit{SC-EffNet\textsubscript{RGB}} among single colorspace models. 

Table \ref{tab:baseline_comparison} presents a comparison of model performances in terms of test accuracies of individual classes and total accuracy for our \textit{MC-EffNet-2} model against the chosen baselines. The results indicate that our model obtains the highest overall accuracy, and highest accuracies even for the individual classes. Among the baselines, the next higher accuracy shown by InceptionResnet is less than our model by 9.92 percentage points. All the models including those proposed for \textit{natural images versus computer graphics} problem, except the model proposed by Quan et al. \cite{quan2018distinguishing}, show a similar trend of higher accuracy for class GAN followed by class Real and then Graphics. Whereas, the model proposed by Quan et al. \cite {quan2018distinguishing} shows higher accuracy for the class Graphics for which the model was originally proposed, but not higher than the accuracy for class Graphics achieved by our \textit{MC-EffNet-2}. 

\setlength{\tabcolsep}{5pt}
\begin{table}[!t]
\centering
\caption{Comparison of model performance accuracies in percentage \\(The highest accuracy is given in boldface)}
\label{tab:baseline_comparison}
\begin{tabular}{lcccc}
\hline
\multicolumn{1}{c}{Model} & GAN & Graphics & Real & Total Accuracy \\
\hline
Quan et al. \cite{quan2018distinguishing} & 56.08 & 72.88 & 69.79 & 66.25 \\
Rezende et al. \cite{de2018exposing} & 59.40 & 52.83 & 57.40 & 56.54 \\
Nataraj et al. \cite{nataraj2019detecting} & 76.00 & 48.25 & 57.13 & 60.46 \\
InceptionResNet \cite{szegedy2017inception} & 85.25 & 73.50 & 79.63 & 79.46 \\
\textit{MC-EffNet-2 (Our model)} & \textbf{96.75} & \textbf{83.75} & \textbf{87.63} & \textbf{89.38}\\
\hline
\end{tabular}
\end{table}

\newcommand{\Rlogo}{\protect\includegraphics[height=1.8ex,keepaspectratio]{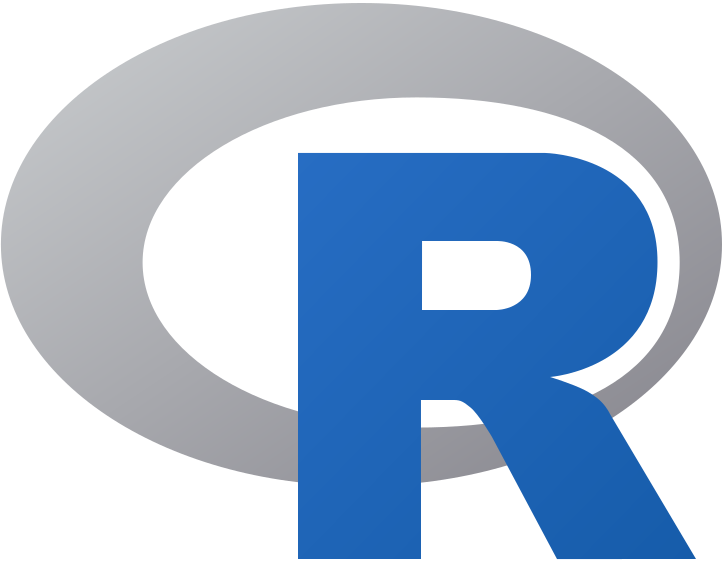}}

\subsubsection{Statistical significance}

In addition to the significant gains achieved by our \textit{MC-EffNet-2} model in terms of accuracy over various baselines, we conduct statistical significance test between our model and the baselines. We perform the Stuart-Maxwell\footnote{\url{http://www.john-uebersax.com/stat/mcnemar.htm\#stuart}} test  
with conventional significance level, i.e., a p-value of 0.05. We obtain a p-value of 0.00187 between our \textit{MC-EffNet-2} model and the model that obtained highest accuracy among the baselines (InceptionResNet), and a p-value of 0.01036 between our \textit{MC-EffNet-2} model and the model that obtained highest accuracy among the Single Colorspace EfficientNet Networks (\textit{SC-EffNet\textsubscript{RGB}}), which provides enough evidence to conclude that the results of our \textit{MC-EffNet-2} model are statistically significant over the best baselines.

\subsubsection{Robustness against Post-processing}

\begin{figure}[!t]
\centering
\includegraphics[width=\linewidth]{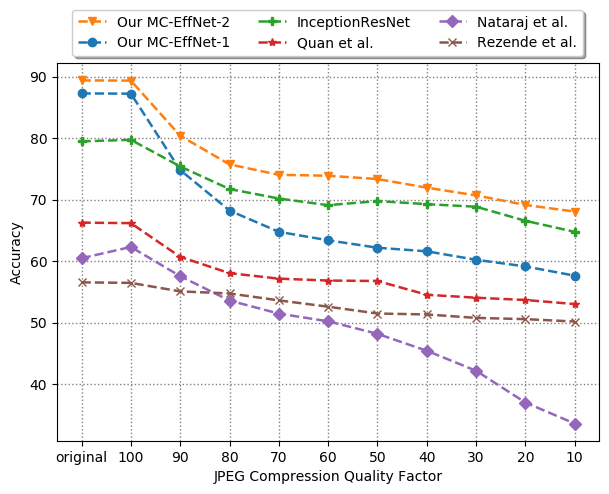}
\caption{Classification accuracies for various JPEG compression quality factors}
\label{fig:robustness}
\end{figure}

Post-processing operations are quite common when uploading images to the web or social media. Therefore, apart from producing good accuracies on original images in the dataset, an effective algorithm for image forensics should also be robust over post-processing operations. We evaluate robustness of our model and baselines towards the typical post-processing operation of JPEG compression, where the models trained on original data are tested over ten different JPEG compression quality factors within the range 100 to 10, in steps of 10. The result of our robustness test is shown in figure \ref{fig:robustness}, where we can observe that even though accuracy of our \textit{MC-EffNet-2} model drops with the decrease of quality factor, it always achieves better performance than the baselines for all the quality factors. Also, we can observe that our \textit{MC-EffNet-2} model attains highly improved robustness than \textit{MC-EffNet-1} with the inclusion of pre-processing block to the colorspace transformations. As like the classification results of original images without compression here also InceptionResnet is the baseline that shows next higher results for different quality factors. The entire results over different compression quality factors indicate that our \textit{MC-EffNet-2} model achieves better robustness towards post-processing based on JPEG compression than the baselines. 

\subsubsection{Generalizability}

We analyze generalizability of our \textit{MC-EffNet-2} model and the baselines by testing over three dataset combinations that are unseen during the training phase. In the first dataset, images for Real and Graphics classes are collected from PIM-Google (photographic images from Google Image Search) and PRCG (photo-realistic computer graphics images) sets of Columbia 
dataset \cite{ng2005columbia}, respectively, and images for class GAN is collected from the generated images of PG\textsuperscript{2} (Pose Guided Person Generation \cite{ma2017pose}) GAN algorithm that produces high quality and realistic person images. The Columbia 
dataset \cite{ng2005columbia} contains large diversity in image content. Many previous state-of-the-art works utilize this dataset for the two-class \textit{natural images versus computer graphics} problem \cite{ng2005physics,quan2018distinguishing,cui2018identifying,ni2019evaluation}. From this dataset with 800 images per class, we collect only twenty percent of the data, i.e., 160 images per class randomly (without considering the five images which were associated with incorrect labels in PIM-Google class as per the findings in \cite{quan2018distinguishing}), since it is only for testing model generalizability. To build a balanced dataset we follow the same number of 160 random images from PG\textsuperscript{2} for class GAN. Apart from PIM-Google and PRCG, Columbia 
dataset also consists of another set, PIM-Personal (photographic images from the authors' personal collections), with 800 natural images which constitutes the second dataset for generalizability test, by replacing class Real of the first dataset with 160 images randomly collected from PIM-Personal class. 

Another dataset combination mostly used in many previous state-of-the-art works \cite{rahmouni2017distinguishing,quan2018distinguishing,ni2019evaluation} for the \textit{natural images versus computer graphics} problem is the RAISE \cite{dang2015raise} versus Level-Design Reference Database \cite{piaskiewicz2017level}. We collected 160 images randomly from RAISE dataset that consists of high-resolution raw and uncompressed images specifically for image forensic research investigations and converted them directly to JPEG format to form class Real of the third dataset. The Level-Design Reference Database consisting of screenshots from various video games can be seen utilized in \cite{rahmouni2017distinguishing} by selecting only those screenshots which seem to be photo-realistic followed by cropping them to remove the gaming information like dialogues, text bars, etc. From these images provided by \cite{rahmouni2017distinguishing}, we collect 160 images to form class Graphics of the third dataset. To form class GAN we collect images generated by CycleGAN \cite{zhu2017unpaired} that generates high-quality realistic images, utilized in many state-of-the-art works to detect GAN images \cite{nataraj2019detecting,marra2019incremental,goebel2020detection}. For CycleGAN images, instead of choosing a single category of images from various unpaired image-to-image translation categories of objects and scenes, we collect 160 images randomly from the horses-to-zebra, zebra-to-horse, apple-to-orange and orange-to-apple categories. Table \ref{tab:generalizability} shows the results of generalizability tests for our model and baselines when tested over the three datasets, where we can clearly observe that our model outperforms all the baselines. Higher accuracies obtained for our model over the datasets on which the model was not originally trained, indicates the promising nature of our approach for tackling future challenges in computer-generated images. Also, we can observe that even though our model is trained on the category of GAN algorithms that generate whole new images, such as StyleGAN, it could perform well on CycleGAN which belongs to the attribute transfer category of GAN algorithms.

\setlength{\tabcolsep}{4.5pt}
\begin{table}[!t]
\centering
\caption{Model generalizability over different datasets in percentage (The highest accuracy is given in boldface)}
\label{tab:generalizability}
\begin{tabular}{lccc}
\hline
\multicolumn{1}{c}{Model} & 
\begin{tabular}[c]{@{}c@{}}PG\textsuperscript{2}$\times$\\PRCG$\times$\\ PIM-Google\end{tabular} & \begin{tabular}[c]{@{}c@{}}PG\textsuperscript{2}$\times$\\PRCG$\times$\\ PIM-Personal\end{tabular} & 
\begin{tabular}[c]{@{}c@{}}Cycle GAN$\times$\\Raise$\times$\\Level-Design\end{tabular} \\ \hline
Quan et al. \cite{quan2018distinguishing} & 54.22 & 56.27 & 60.01 \\
Rezende et al. \cite{de2018exposing} & 51.25 & 51.21 & 50.92 \\
Nataraj et al. \cite{nataraj2019detecting} & 49.17 & 53.63 & 48.75 \\
InceptionResNet \cite{szegedy2017inception} & 62.08 & 67.16 & 71.74 \\
\textit{MC-EffNet-2 (Our model)} & \textbf{81.04} & \textbf{85.21} & \textbf{84.79} \\
\hline
\end{tabular}
\end{table}

\subsubsection{Feature Visualization}

\begin{figure*}[!t]
\centering
\fbox{\subfloat[Raw image features]{\includegraphics[width=0.48\textwidth]{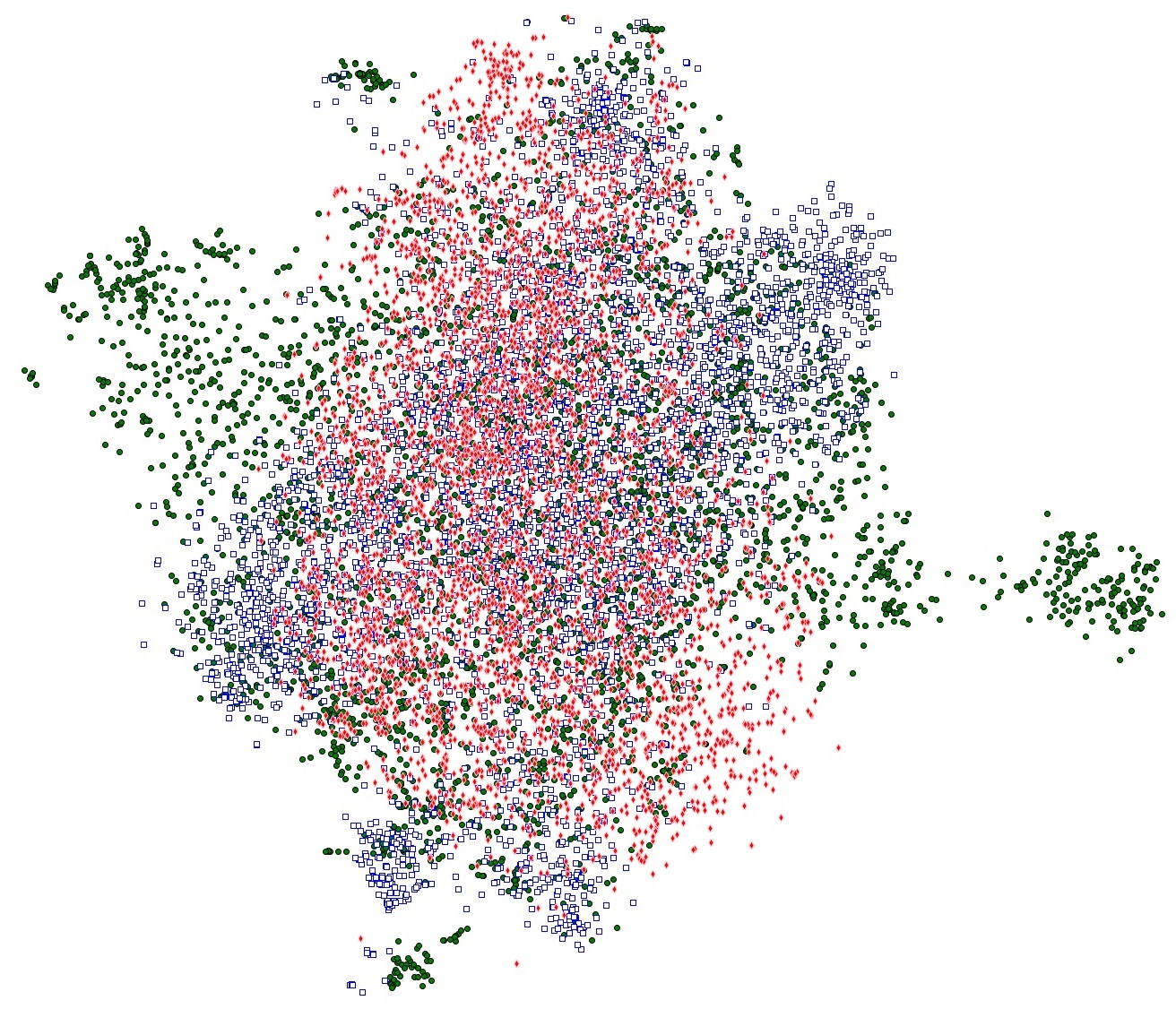}}%
\label{fig:rgb_viz}}
\hfil
\fbox{\subfloat[\textit{MC-EffNet-2} features]{\includegraphics[width=0.48\textwidth]{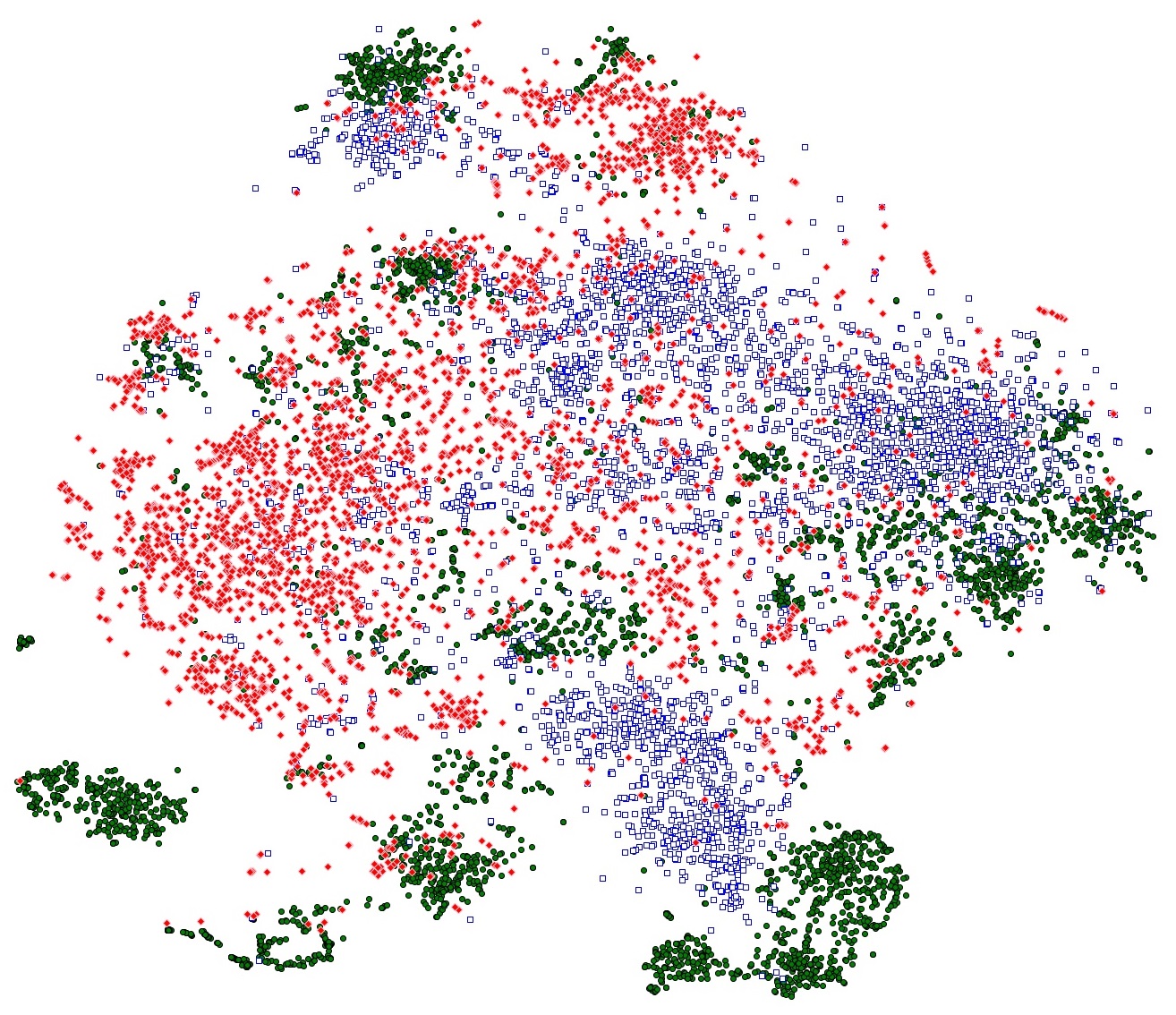}}%
\label{fig:mc_effnet2_viz}}
\caption{t-SNE visualizations of the feature vectors. (\tikzsymbol{fill=cadmiumgreen} indicates GAN, \tikzsymbol[rectangle]{minimum width=6pt,minimum height=6pt,blue,fill=white} indicates Graphics and, \tikzsymbol[diamond]{pastelpink,fill=mediumcandyapplered,} indicates Real images)}
\label{fig:tsne}
\end{figure*}

Our \textit{MC-EffNet-2} model, projects raw pixels of input images with dimension 224$\times$224$\times$3, or feature vector of size 150528  to a lower dimension feature vector of size 3840, with an intention to provide good amount of separability between the three classes so that the top classifier layer attains a high classification accuracy. To understand the separability of features projected from our model we implement a technique for dimensionality reduction called t-Distributed Stochastic Neighbor Embedding (t-SNE) \cite{van2008visualizing} that can visualize high dimensional features into a two-dimensional plane. We project both the raw image features, and the output features from \textit{MC-EffNet-2} into two-dimensional plots, with three different colors indicating three different classes. The plots are given in figure \ref{fig:tsne} where green circles represent the class GAN, pink diamonds represent the class Real and blue squares represent the class Graphics. As can be seen from the t-SNE visualizations, the raw image features are more clustered particularly towards the center of the plot, whereas, the output features from our \textit{MC-EffNet-2} model are seen to be more separated. Thus, the t-SNE visualizations prove that our \textit{MC-EffNet-2} model suits the forensic task of classifying natural images from computer-generated images including both computer graphics and GAN images by projecting the raw image pixels to a much better and separable feature space.

\subsubsection{Psychophysics Experiments}

We were also curious about how accurately humans can distinguish natural images from photo-realistic computer-generated images including computer graphics and GAN images. Hence, in this study, we also perform a manual classification test for GAN, Graphics and Real class of images by gathering information from eleven human participants on a set of 330 images randomly selected from the test data used in our study. The participants belong to the age group 22 to 40 years with normal or corrected-to-normal visual acuity and normal color vision. We utilize an annotation tool VIA \cite{dutta2016via}, that helps participants to label each image as GAN, Graphics or Real, and also to mark parts/regions in the image that explains their decisions. This way of asking human participants to provide evidence/explanation in the form of region markings allows to omit the chances of lucky guesses and moreover provides insight to what the participants perceive as a suspect and/or evidence in the images. 
In VIA the participants can zoom and or zoom out images for better analysis and no other constraints like viewing distance, time for observation, etc. are imposed. Each participant is asked to label thirty images randomly chosen and assigned to them and each image is annotated only once. For a participant, the entire experiment on thirty images including marking their explanations in the images takes nearly 45 minutes. 

\begin{figure}[!t]
\centering
\subfloat[Manual classification]{\includegraphics[width=0.33\textwidth]{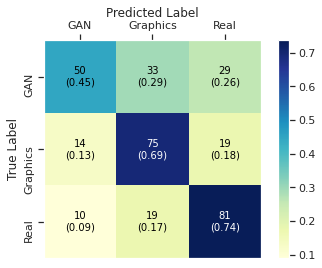}%
\label{fig:psycho_manual}}
\hfil
\subfloat[\textit{MC-EffNet-2}]{\includegraphics[width=0.33\textwidth]{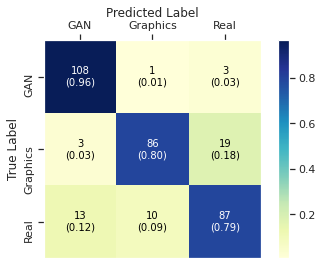}%
\label{fig:psycho_mceffnet2}}
\caption{Confusion matrices of classification performed by human participants and our \textit{MC-EffNet-2} model}
\label{fig:psycho_acc}
\end{figure}

We also compute the accuracy of our \textit{MC-EffNet-2} model over the same set of 330 images selected for psychophysics experiments, for comparison. Figure \ref{fig:psycho_acc} shows the confusion matrices of manual classification performed by human participants and that of our \textit{MC-EffNet-2} model over 330 images. For manual classification, we obtain a total accuracy of 62.42 percentage, whereas for the same set of images our \textit{MC-EffNet-2} model obtains a higher accuracy of 85.15 percentage, i.e., a very high gain of 22.73 percentage points. We can observe that the ability of humans to classify Real images is almost near to our \textit{MC-EffNet-2} model with a decrease of 5 percentage points for manual classification, but for Graphics images \textit{MC-EffNet2} highly outperforms manual classification. Similarly, in case of GAN images, manual classification accuracy is almost half of \textit{MC-EffNet-2}. The overall results indicate that the ability of humans to identify photo-realistic computer-generated images is very low and hence there is a high necessity of image forensic algorithms that can computationally aid to distinguish natural images and photo-realistic computer-generated images. Our \textit{MC-EffNet-2} model with a high classification accuracy, especially for the photo-realistic computer-generated images, is thus a better solution for the forensic task of distinguishing natural images from photo-realistic computer-generated images. We also compute the Stuart-Maxwell statistical significance test between our model and manual classifcation, where we obatin a p-value of 2.481E-06 indicating the significance of our \textit{MC-EffNet-2} model over manual classification.

\subsection{Understanding the Explanations}

Apart from analyzing model performance, we also investigate the behavior of our \textit{MC-EffNet-2} model so that we can trust the predictions of our deep learning model. We utilize Gradient-weighted Class Activation Mapping (Grad-CAM) \cite{selvaraju2017grad} that makes use of class-specific gradient information to make a deep learning model more transparent through visual explanations. To obtain visual explanations we employ Grad-CAM at the penultimate layer of our fully trained and saved\textit{ MC-EffNet-2} model, which constructs coarse localization maps of the salient regions in input images that are significantly important for the predictions. 

Since our task is formulated as an image classification problem build using transfer learning methodology over the source networks of pre-trained EfficientNets, originally meant for object classification task, first, we try to identify whether our fused model is still looking for objects while making the decisions, or is it looking for the regions significant for classifying natural images and computer-generated ones in a forensic perspective. To address this question we take the Grad-CAM explanations of our \textit{MC-EffNet-2} model and also the base network, EfficientNetB0 which is pre-trained on the ImageNet data, on a set of images in our dataset. The Grad-CAM explanations from both the models for GAN, Graphics and Real images are shown in table \ref{tab:gradecam_comparison}. The EfficientNetB0 network which is pre-trained on the ImageNet data, as obvious, mainly highlights the objects present in images. But, we can observe that, even though our entire fused \textit{MC-EffNet-2} model is built over the base network of EfficientNet which was originally designed for an object classification task, after applying transfer learning towards our forensic task, does not primarily give importance to the objects as in the source task, rather highlights the regions that are significant in classifying them as GAN, Graphics or Real in a forensic perspective. This indicates the fitness of our \textit{MC-EffNet-2} model as a forensic solution to classify GAN, Graphics and Real images.

\setlength{\tabcolsep}{1pt}
\begin{table*}[!t]
\centering
\caption{Grad-CAM explanations from the base network EfficientNetB0 and our \textit{MC-EffNet-2} model for GAN, Graphics and Real images}
\label{tab:gradecam_comparison}
\begin{tabular}{lll||lll||lll}
\hline
\multicolumn{3}{c}{GAN} & \multicolumn{3}{c}{Graphics} & \multicolumn{3}{c}{Real}
\\ \hline
\multicolumn{1}{c}{\multirow{2}{*}{\begin{tabular}[c]{@{}c@{}}Original \\ image\end{tabular}}} & \multicolumn{2}{c||}{Grad-CAM explainations} & \multicolumn{1}{c}{\multirow{2}{*}{\begin{tabular}[c]{@{}c@{}}Original \\ image\end{tabular}}} & \multicolumn{2}{c||}{Grad-CAM explainations} & \multicolumn{1}{c}{\multirow{2}{*}{\begin{tabular}[c]{@{}c@{}}Original \\ image\end{tabular}}} & \multicolumn{2}{c}{Grad-CAM explainations} \\ \cline{2-3} \cline{5-6} \cline{8-9} 
\multicolumn{1}{c}{} & \multicolumn{1}{c}{EfficientNetB0} & \multicolumn{1}{c||}{\textit{MC-EffNet-2}} & \multicolumn{1}{c}{} & \multicolumn{1}{c}{EfficientNetB0} & \multicolumn{1}{c||}{\textit{MC-EffNet-2}} & \multicolumn{1}{c}{} & \multicolumn{1}{c}{EfficientNetB0} & \multicolumn{1}{c}{\textit{MC-EffNet-2}} \\ \hline
\includegraphics[width=19mm, height=19mm, valign=c]{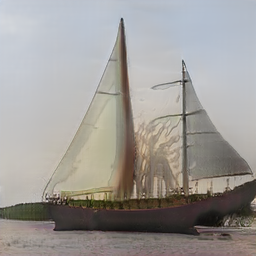} 
& \includegraphics[width=19mm, height=19mm, valign=c]{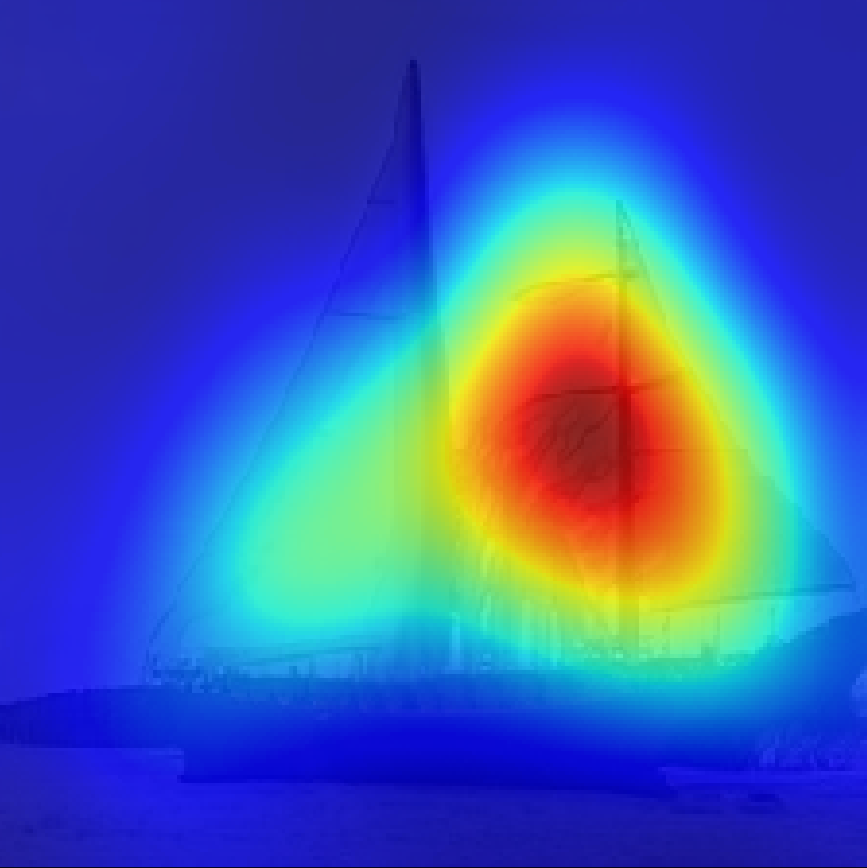}
& \includegraphics[width=19mm, height=19mm, valign=c]{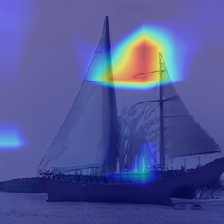}
& \includegraphics[width=19mm, height=19mm, valign=c]{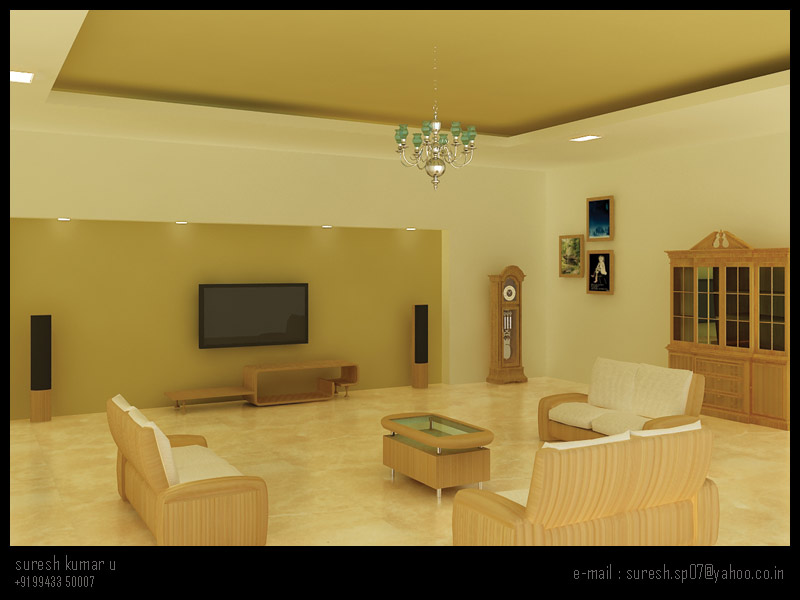}  
& \includegraphics[width=19mm, height=19mm, valign=c]{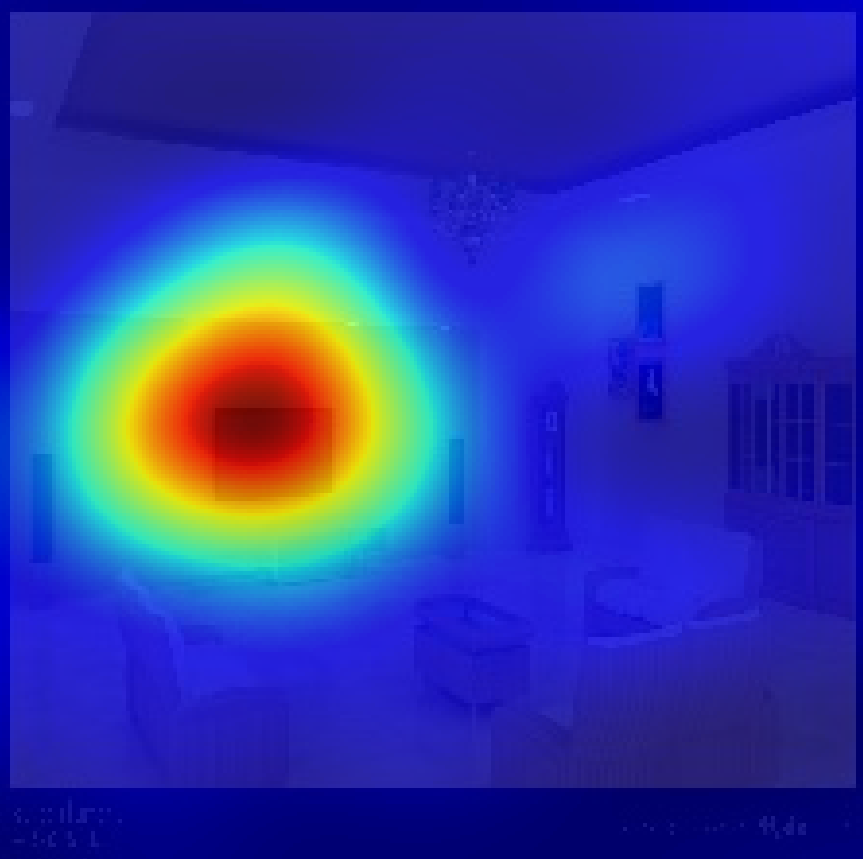} 
& \includegraphics[width=19mm, height=19mm, valign=c]{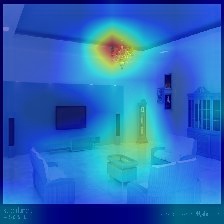}
& \includegraphics[width=19mm, height=19mm, valign=c]{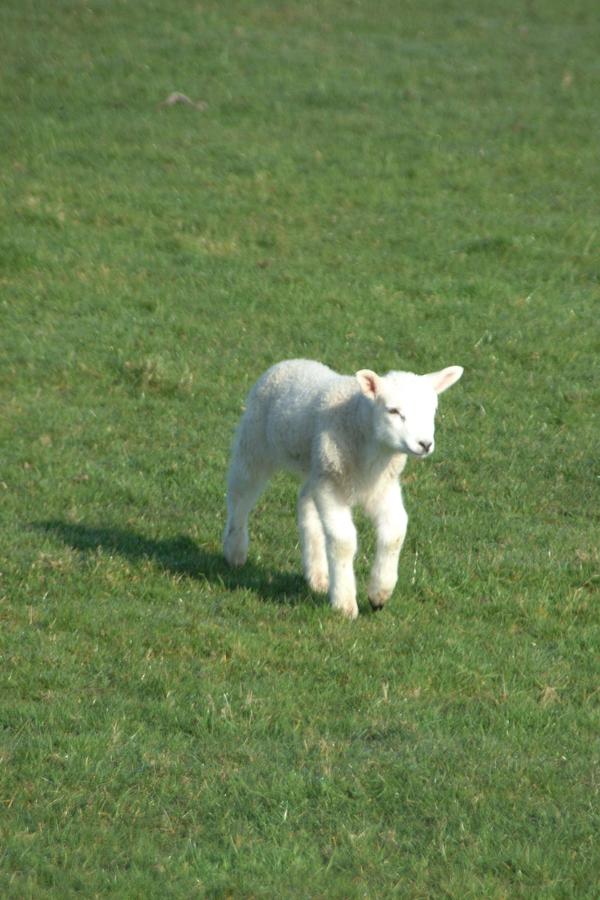} 
& \includegraphics[width=19mm, height=19mm, valign=c]{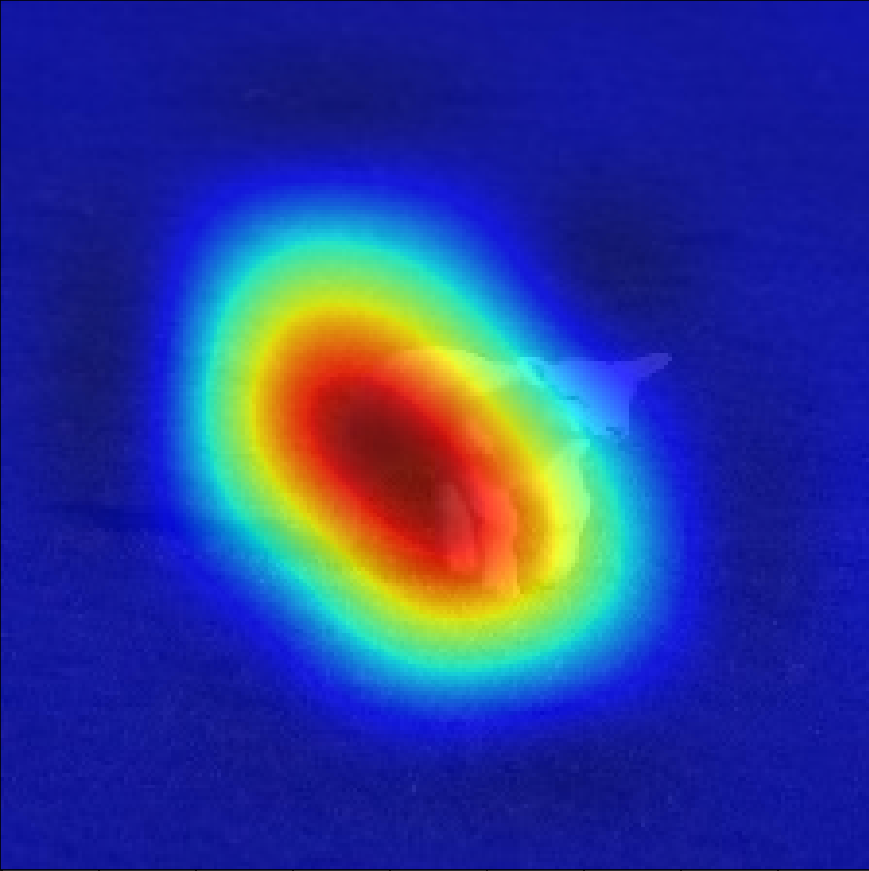} & \includegraphics[width=19mm, height=19mm, valign=c]{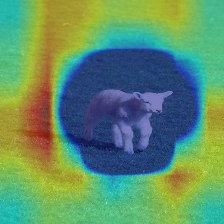}  \\
\includegraphics[width=19mm, height=19mm, valign=c]{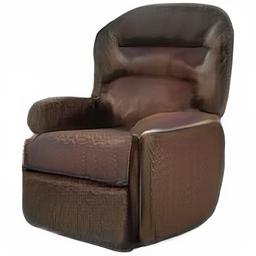} 
& \includegraphics[width=19mm, height=19mm, valign=c]{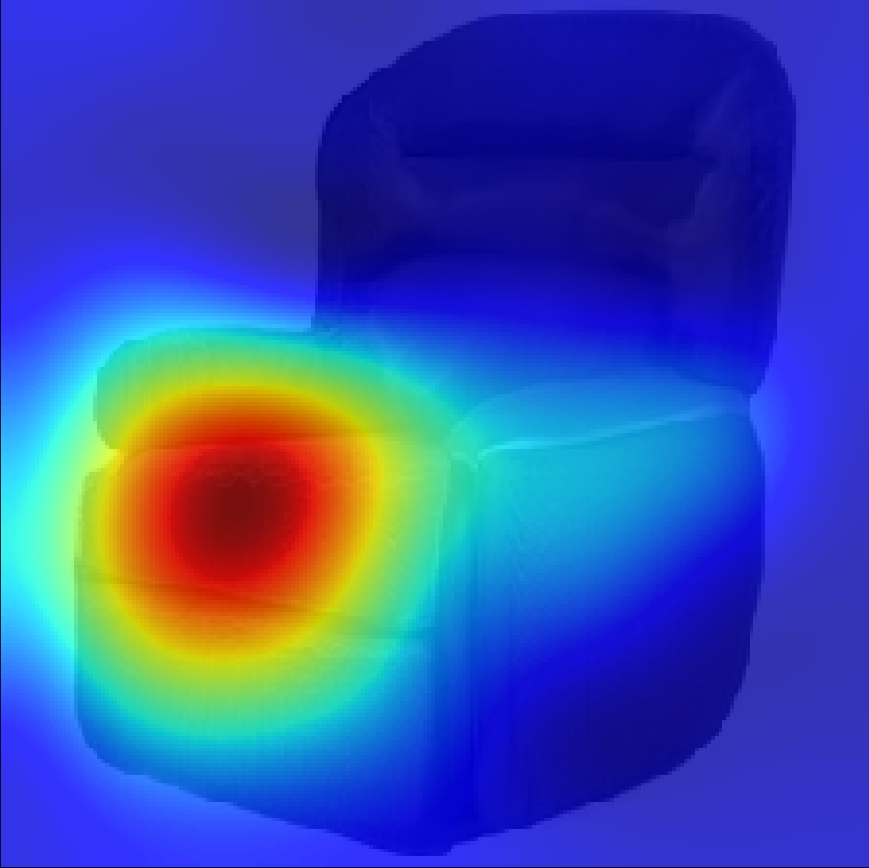} 
& \includegraphics[width=19mm, height=19mm, valign=c]{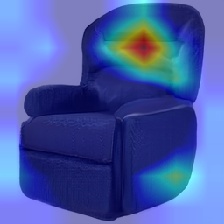}
& \includegraphics[width=19mm, height=19mm, valign=c]{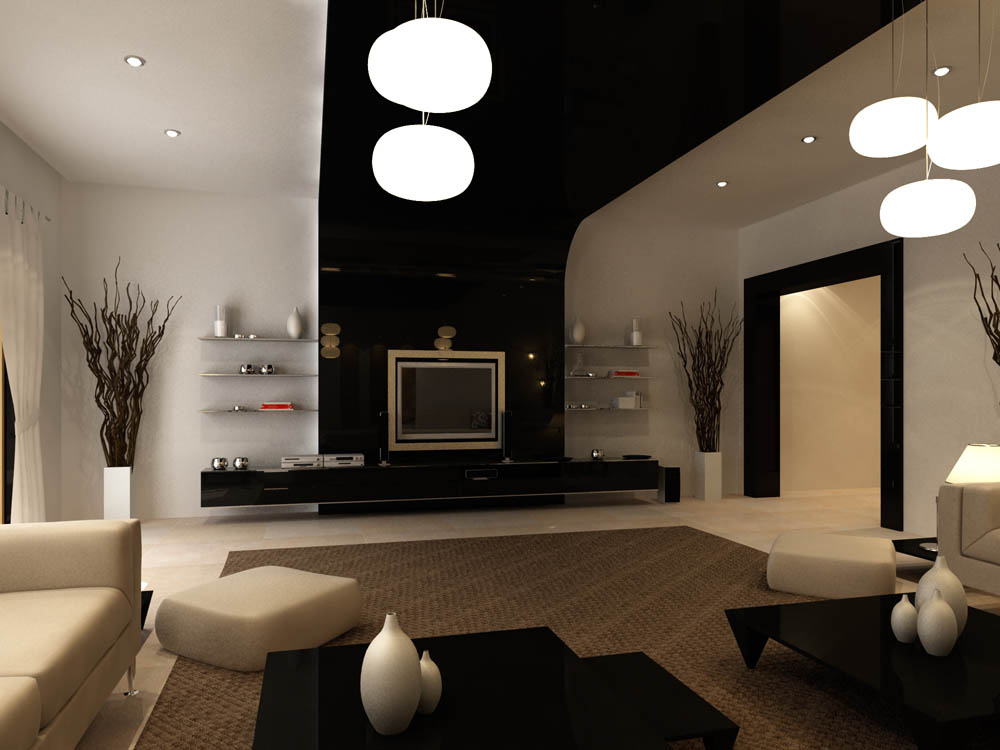}  
& \includegraphics[width=19mm, height=19mm, valign=c]{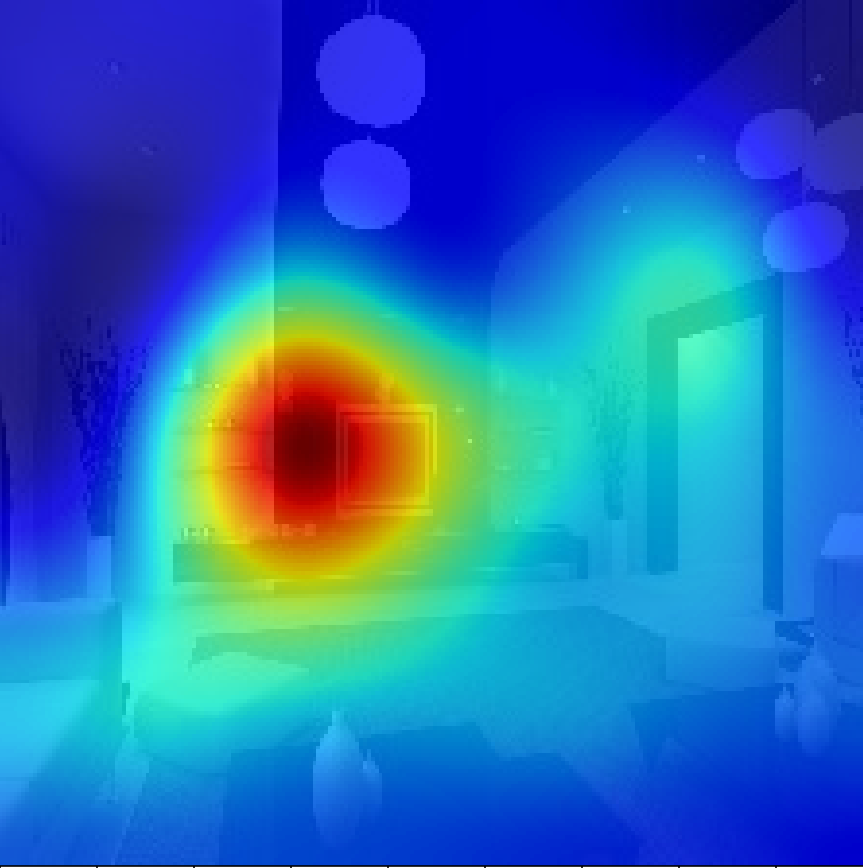}  
& \includegraphics[width=19mm, height=19mm, valign=c]{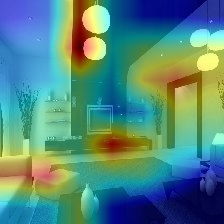}
& \includegraphics[width=19mm, height=19mm, valign=c]{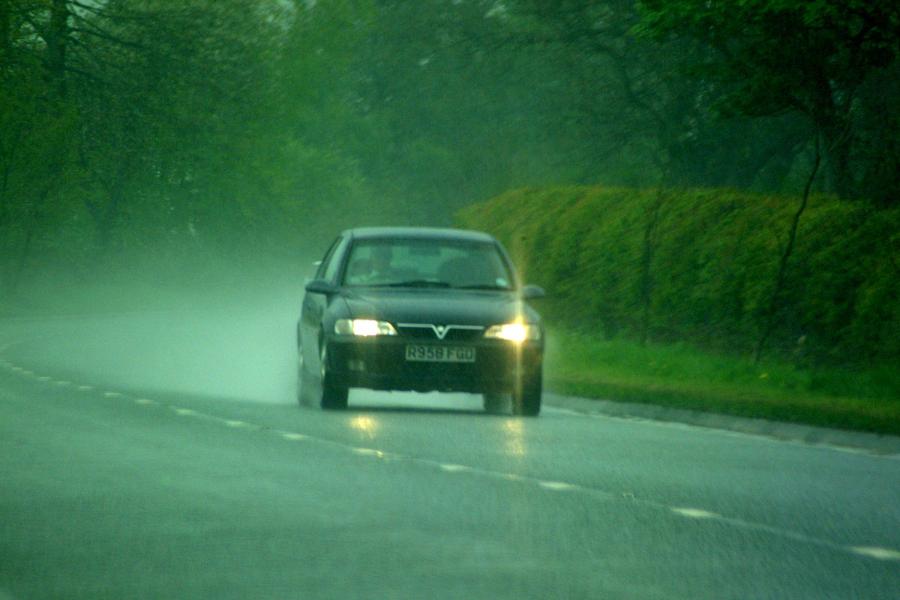}
& \includegraphics[width=19mm, height=19mm, valign=c]{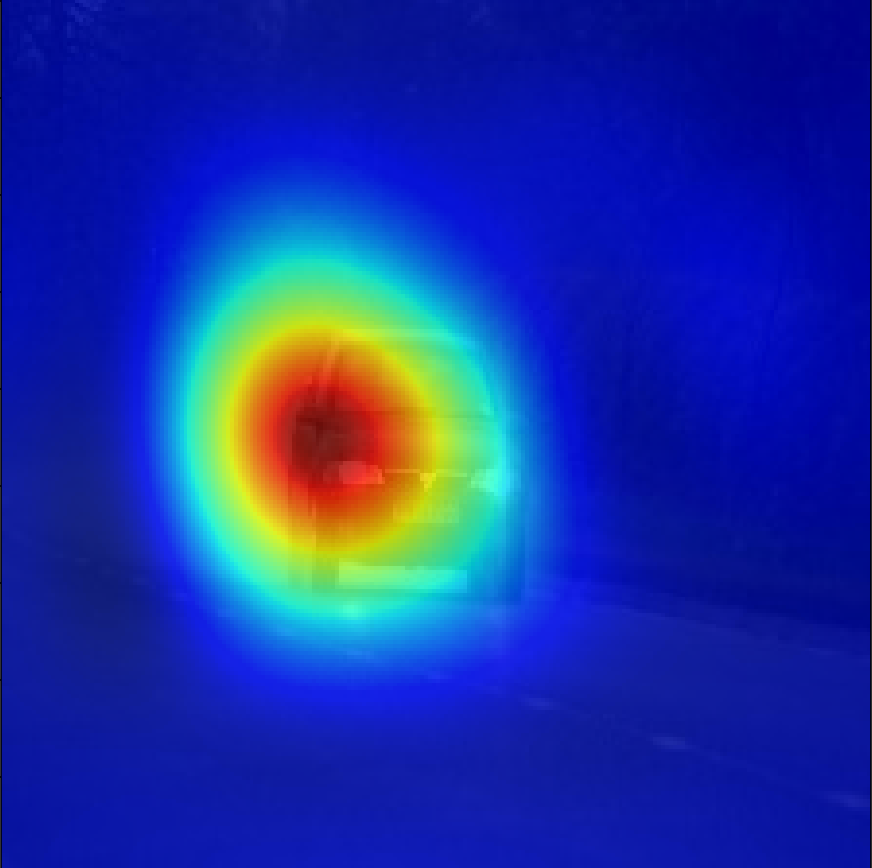}
& \includegraphics[width=19mm, height=19mm, valign=c]{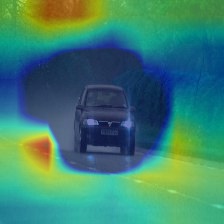}  \\
\hline
\end{tabular}
\end{table*}

Accordingly, we try to understand what makes our \textit{MC-EffNet-2} model label an image as GAN, Graphics or Real in context of image forensics. In this stage, we also utilize the manual explanations in the form of region markings (yellow bounding boxes\footnote{The bounding box boundaries are thickened for better visibility}) from the human participants of our psychophysics experiments, to compare whether the Grad-CAM explanations given by our model have any similarities with manual explanations. We first take the case of images for which our model and human participants provide correct predictions and analyze the Grad-CAM and manual explanations of these images. Table \ref{tab:correct_explanations} shows the examples of this case, where each set of three columns indicates the GAN, Graphics and Real classes, respectively, and for each class, we provide three sample images with their corresponding Grad-CAM and manual explanations. In the first image of a sheep among the GAN generated ones, we can observe that the image is not completely formed and there are missing regions like legs of the sheep. When we look at the Grad-CAM explanations provided by our \textit{MC-EffNet-2} model, we can see that the model captures image regions of legs and also the slight difference in fur color and texture near the neck region. For the same image, human participant has marked the leg region as an explanation for their decision of labeling that image as GAN image. Similarly in the second GAN image of a bird where the image is not completely formed at the head region and the texture of bird is misplaced towards bottom of the image, Grad-CAM captures both these regions along with the tail region of bird and also some part of the surroundings. The human participant has manually marked the region at head and misplaced texture at the bottom of the image. In next GAN image of a human face where no misformations or misplacements can be seen evidently, Grad-CAM mainly highlights texture of hair, some regions on face and also some part of the surroundings. In this case, human participant has highlighted the hair region as an explanation. Hence, in the GAN image examples, we can see that there are similarities between Grad-CAM explanations provided by our model and manual explanations. 

Next, we analyze Grad-CAM and manual explanations of Graphics images in the second set of columns. For all the correct predictions of Graphics images Grad-CAM explanations are most commonly seen to highlight uneven illuminations or illuminated regions in the images. Human participants are also seen to mark such regions of uneven illuminations. In the third set of Real images, Grad-CAM explanations are commonly seen to be centered on the surroundings, focussing on complex variabilities in the background regions. In the first Real image, along with the surrounding regions a major significance can be seen given to the shadow of swan in water, by Grad-CAM as well as the human participant. Similarly, the feather regions in second image and the clouds in third image can be seen highlighted in both Grad-CAM and manual explanations. From the overall results, we can sum up that the explanations of our \textit{MC-EffNet-2} model are mostly similar to manual explanations, and also our model is able to identify more number of salient regions than human participants, to distinguish natural and computer-generated images. The explanations demonstrate the powerful nature of our \textit{MC-EffNet-2} model to take the decisions meaningfully.

\setlength{\tabcolsep}{1pt}
\begin{table*}[!t]
\centering
\caption{Explanations of images for which our \textit{MC-EffNet-2} model and human participants both produces correct predictions}
\label{tab:correct_explanations}
\begin{tabular}{ccc||ccc||ccc}
\hline
\multicolumn{3}{c}{GAN} & \multicolumn{3}{c}{Graphics} & \multicolumn{3}{c}{Real} \\
\hline
\begin{tabular}[c]{@{}c@{}}Original\\image\end{tabular}
&\begin{tabular}[c]{@{}c@{}}Grad-CAM\\explaination\end{tabular}
&\begin{tabular}[c]{@{}c@{}}Manual\\explanation\end{tabular} 
&\begin{tabular}[c]{@{}c@{}}Original\\image\end{tabular}
&\begin{tabular}[c]{@{}c@{}}Grad-CAM\\explaination\end{tabular}
&\begin{tabular}[c]{@{}c@{}}Manual\\explanation\end{tabular}  &\begin{tabular}[c]{@{}c@{}}Original\\image\end{tabular}
&\begin{tabular}[c]{@{}c@{}}Grad-CAM\\explaination\end{tabular}
&\begin{tabular}[c]{@{}c@{}}Manual\\explanation\end{tabular} 
\\ \hline \vspace{2pt}
\includegraphics[width=19mm, height=19mm, valign=c]{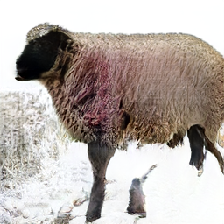}
& \includegraphics[width=19mm, height=19mm, valign=c]{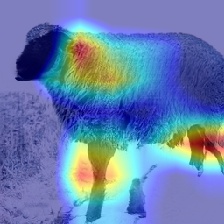}
& \includegraphics[width=19mm, height=19mm, valign=c]{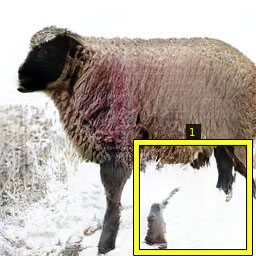}
&  \includegraphics[width=19mm, height=19mm, valign=c]{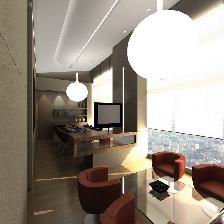}
& \includegraphics[width=19mm, height=19mm, valign=c]{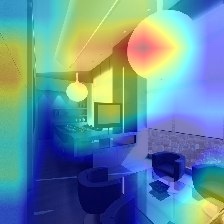}
& \includegraphics[width=19mm, height=19mm, valign=c]{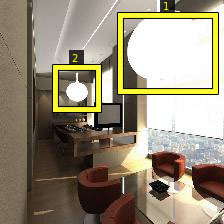}
& \includegraphics[width=19mm, height=19mm, valign=c]{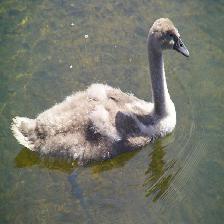}
& \includegraphics[width=19mm, height=19mm, valign=c]{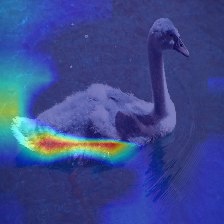}
& \includegraphics[width=19mm, height=19mm, valign=c]{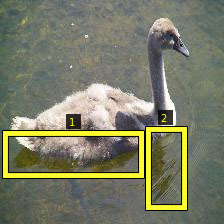}
\\ \vspace{2pt}
\includegraphics[width=19mm, height=19mm, valign=c]{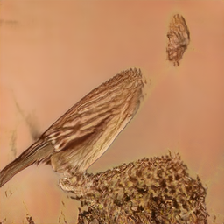}
& \includegraphics[width=19mm, height=19mm, valign=c]{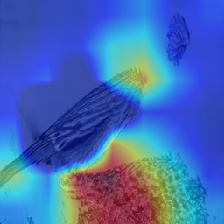}
& \includegraphics[width=19mm, height=19mm, valign=c]{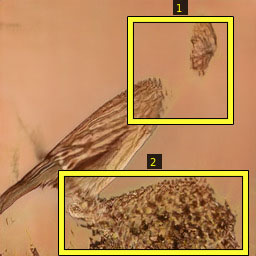}
& \includegraphics[width=19mm, height=19mm, valign=c]{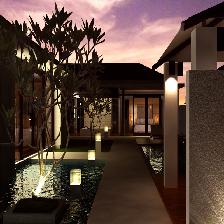}
& \includegraphics[width=19mm, height=19mm, valign=c]{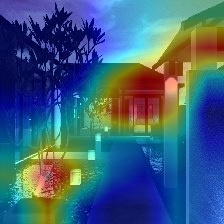}
& \includegraphics[width=19mm, height=19mm, valign=c]{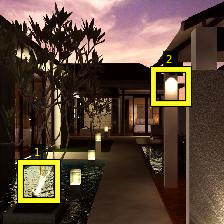}
&  \includegraphics[width=19mm, height=19mm, valign=c]{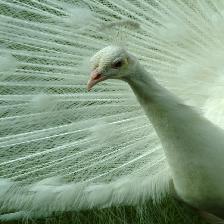}
& \includegraphics[width=19mm, height=19mm, valign=c]{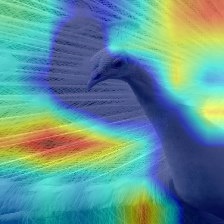}
&  \includegraphics[width=19mm, height=19mm, valign=c]{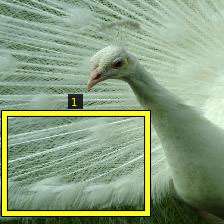}
\\ \vspace{2pt}
\includegraphics[width=19mm, height=19mm, valign=c]{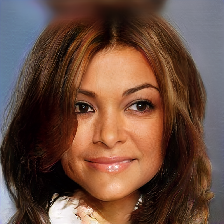}
& \includegraphics[width=19mm, height=19mm, valign=c]{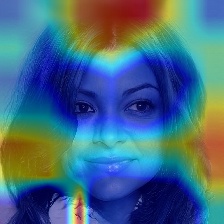}
& \includegraphics[width=19mm, height=19mm, valign=c]{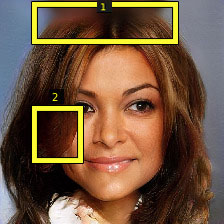}
& \includegraphics[width=19mm, height=19mm, valign=c]{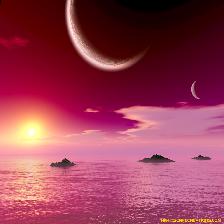}
& \includegraphics[width=19mm, height=19mm, valign=c]{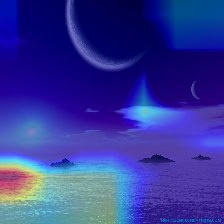}
& \includegraphics[width=19mm, height=19mm, valign=c]{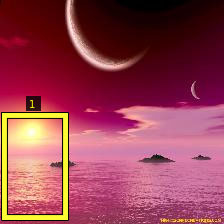}
& \includegraphics[width=19mm, height=19mm, valign=c]{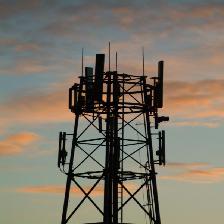}
& \includegraphics[width=19mm, height=19mm, valign=c]{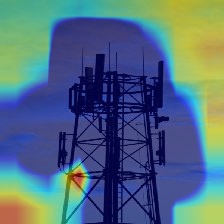}
& \includegraphics[width=19mm, height=19mm, valign=c]{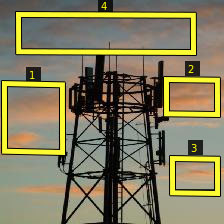}
\\ 
\hline
\end{tabular}
\end{table*}

We also try to get insights on wrong predictions by examining images for which the human participants provide correct predictions but our \textit{MC-EffNet-2} model produces wrong predictions. A few examples of this case are given in table \ref{tab:wrong_explanations}. The first image (a) is GAN which is manually predicted as GAN itself but \textit{MC-EffNet-2} misclassifies it as a Real. The Grad-CAM explanation from \textit{MC-EffNet-2} shows that the decision is produced from the head region, rock at bottom of the image and the surroundings but not mainly from the misformed regions like the leg region. Similarly in (d), Graphics is misclassified as Real taking into account the region of mountains and some parts of the background rather than considering the regions of illuminations in the image as usually seen in case of graphics images. For Graphics image (c) which is manually predicted as a Graphics itself, \textit{MC-EffNet-2} misclassifies it as a GAN. The Grad-CAM explanation shows that it produces decision from only the object region of one of the photographs in the image and not the regions of illuminations as usually seen in case of Graphics images. Similarly, in Real image (e), the object regions of sheeps are highlighted by \textit{MC-EffNet-2} than the meaningful explanations like shadows or surroundings as usually seen for the Real images, which might be the reason for its misclassification as GAN. Whereas, in the images (b,f) \textit{MC-EffNet-2} highlights the uneven illuminations which might be the reason for its misclassification as Graphics image.

\setlength{\tabcolsep}{1pt}
\begin{table*}[!t]
\centering
\caption{Explanations of images for which human participants provide correct predictions but \textit{MC-EffNet-2} produces wrong predictions}
\label{tab:wrong_explanations}
\begin{tabular}{ccc||ccc||ccc}
\hline
\multicolumn{3}{c}{GAN} & \multicolumn{3}{c}{Graphics} & \multicolumn{3}{c}{Real} \\
\hline
\begin{tabular}[c]{@{}c@{}}Original\\image\end{tabular}
&\begin{tabular}[c]{@{}c@{}}Grad-CAM\\explaination\end{tabular}
&\begin{tabular}[c]{@{}c@{}}Manual\\explanation\end{tabular} 
&\begin{tabular}[c]{@{}c@{}}Original\\image\end{tabular}
&\begin{tabular}[c]{@{}c@{}}Grad-CAM\\explaination\end{tabular}
&\begin{tabular}[c]{@{}c@{}}Manual\\explanation\end{tabular}  &\begin{tabular}[c]{@{}c@{}}Original\\image\end{tabular}
&\begin{tabular}[c]{@{}c@{}}Grad-CAM\\explaination\end{tabular}
&\begin{tabular}[c]{@{}c@{}}Manual\\explanation\end{tabular} 
\\ \hline \vspace{2pt}
\includegraphics[width=19mm, height=19mm, valign=c]{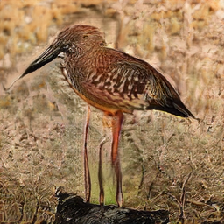}
& \includegraphics[width=19mm, height=19mm, valign=c]{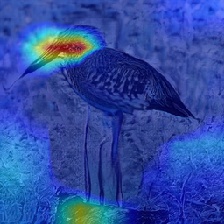}
&  \includegraphics[width=19mm, height=19mm, valign=c]{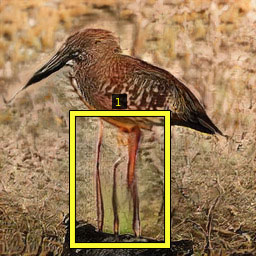}
& \includegraphics[width=19mm, height=19mm, valign=c]{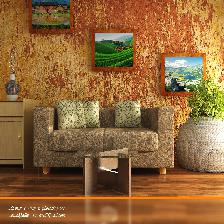}
& \includegraphics[width=19mm, height=19mm, valign=c]{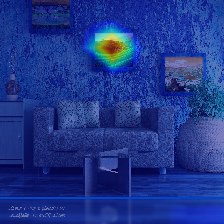}
&  \includegraphics[width=19mm, height=19mm, valign=c]{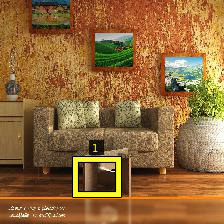}
& \includegraphics[width=19mm, height=19mm, valign=c]{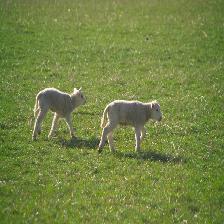}
& \includegraphics[width=19mm, height=19mm, valign=c]{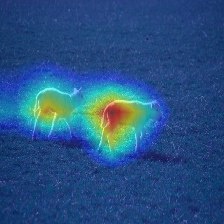}
& \includegraphics[width=19mm, height=19mm, valign=c]{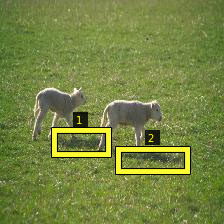}
\\ 
\multicolumn{3}{c||}{(a) \textit{MC-EffNet-2} miclassifies GAN as Real} 
& \multicolumn{3}{c||}{(c) \textit{MC-EffNet-2} miclassifies Graphics as GAN} 
& \multicolumn{3}{c}{(e) \textit{MC-EffNet-2} miclassifies Real as GAN} 
\\ \hline 
\includegraphics[width=19mm, height=19mm, valign=c]{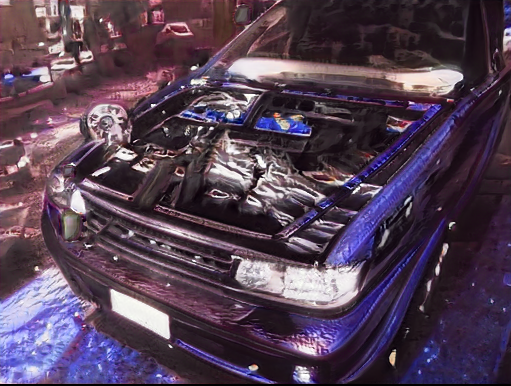}
& \includegraphics[width=19mm, height=19mm, valign=c]{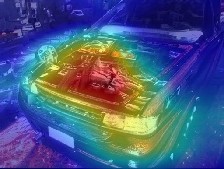}
& \includegraphics[width=19mm, height=19mm, valign=c]{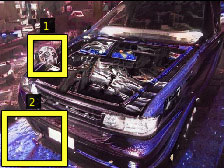}
& \includegraphics[width=19mm, height=19mm, valign=c]{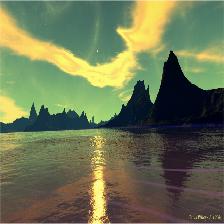}
& \includegraphics[width=19mm, height=19mm, valign=c]{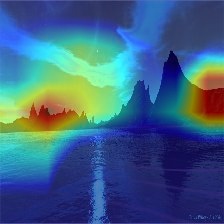}
&  \includegraphics[width=19mm, height=19mm, valign=c]{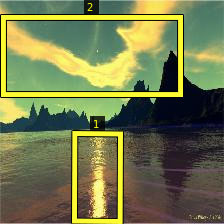}
& \includegraphics[width=19mm, height=19mm, valign=c]{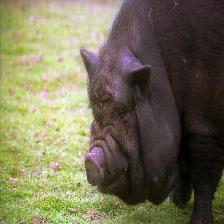}
& \includegraphics[width=19mm, height=19mm, valign=c]{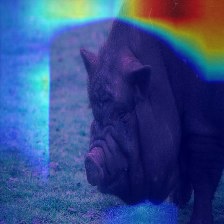}
& \includegraphics[width=19mm, height=19mm, valign=c]{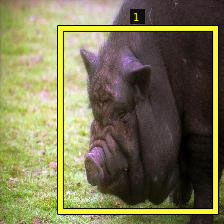}
\\ 
\multicolumn{3}{c||}{(b) \textit{MC-EffNet-2} miclassifies GAN as Graphics} 
& \multicolumn{3}{c||}{(d) \textit{MC-EffNet-2} miclassifies Graphics as Real} 
& \multicolumn{3}{c}{(f) \textit{MC-EffNet-2} miclassifies Real as Graphics} 
\\ \hline 
\end{tabular}
\end{table*}

\section{Conclusion}
\label{sec:conclusion}

In this work, we proposed deep learning based Multi-Colorspace fused EfficientNet model to classify natural images and photo-realistic computer-generated images including both computer graphics and GAN images, as against the state-of-the-art works that have always discussed either \textit{natural images versus computer graphics} or \textit{natural images versus GAN images} problem, at a time. We compared our model with state-of-the-art methods where our model outperforms all the baselines in terms of performance accuracy, robustness against typical post-processing operation of JPEG compression, and generalizability towards other datasets, which demonstrates the utility of our model in real-world forensic applications. We also conducted psychophysics experiments to realize how capable humans are in classifying natural images and photo-realistic computer-generated images, where, the results  of  manual classification accuracy was lower than our model accuracy, particularly in classifying the photo-realistic computer-generated images, indicating the necessity and usefulness of our computational model for the task. We also analyzed the behavior of our model by visualizing the salient regions in the images that are responsible for classification decisions. We compared these visual explanations of our model with the explanations manually labeled by the human participants for their correct predictions, where we could observe similarities between the explanations of our model and manual explanations, indicating that our model takes decisions meaningfully. To our best knowledge, such a comparison of visual explanations to understand whether our model behaves alike human explanations to produce the decisions meaningfuly is a new attempt that might even be useful in other digital image forensics or multimedia security tasks. To aid future research, these manual classifications along with the manually labelled visual explanations,  and other relevant materials, including the source code will be made publicly available at  \url{https://github.com/manjaryp/GANvsGraphicsvsReal} and \url{https://dcs.uoc.ac.in/cida/projects/dif/mceffnet.html} along with the publication. In the future, we are considering to extend our work to identify other forensic attacks like the recaptured images. We are also considering to extend our work to classify natural and computer generated videos.

\section*{Acknowledgment}

The authors would like to thank Prof. Dr. Anderson Rocha from University of Campinas for sharing their dataset in \cite{tokuda2013computer}, the authors of \cite{quan2018distinguishing,de2018exposing,ni2019evaluation} for making thier source codes publicly available, and the authors of \cite{ng2005columbia,rahmouni2017distinguishing} for making thier datasets publicly available,. The authors would also like to thank the Master students (CS2019-21) and staff at the Department of Computer Science, University of Calicut for their involvement and co-operation to conduct the psychophysics experiments in this work.
\ifCLASSOPTIONcaptionsoff
  \newpage
\fi



\bibliographystyle{IEEEtran}
\bibliography{references}
\end{document}